\documentclass[11pt, a4paper]{lumia}

\usepackage[sort&compress]{natbib}
\usepackage{xspace}
\newcommand*{\eg}{\textit{e.g.}\xspace}

\theoremstyle{plain}

\newtheorem*{proposition*}{Proposition}

\theoremstyle{definition}

\theoremstyle{definition}

\def\eqref#1{equation~\ref{#1}}

\usepackage{graphicx}           
\usepackage{tikz}               
\usepackage[edges]{forest}      

\usepackage{url}                
\usepackage{xurl}               

\usepackage{array}              
\usepackage{longtable}          
\usepackage{multirow}           
\usepackage{makecell}           
\usepackage{ragged2e}           

\usepackage{mathtools}          
\usepackage{nicefrac}           

\usepackage{algorithm}          
\usepackage{algorithmic}
\usepackage{cleveref}
\usepackage{listings}           

\usepackage{subcaption}         
\usepackage{wrapfig}            
\usepackage[export]{adjustbox}  

\usepackage{xspace}             
\usepackage[normalem]{ulem}     
\usepackage{CJKutf8}            

\usepackage[tikz]{bclogo}       
\usepackage[framemethod=tikz]{mdframed} 

\usepackage{lipsum}             
\usepackage{tocloft}            
\usepackage{afterpage}          
\usepackage{bbding}             
\usepackage{epigraph}           
\usepackage{minitoc}            
\usepackage{multicol}           
\usepackage{textgreek}          

\newcolumntype{P}[1]{>{\RaggedRight\arraybackslash}p{#1}}

\definecolor{uclablue}{RGB}{39, 116, 174}
\definecolor{bigaired}{RGB}{156, 0, 0}
\definecolor{myblue}{HTML}{598BE7}
\definecolor{mildblue}{RGB}{31,119,180}
\definecolor{sectionblue}{RGB}{70, 130, 180}
\definecolor{methodblue}{RGB}{0, 150, 136}
\definecolor{bgblue}{RGB}{245,243,253}
\definecolor{ttblue}{RGB}{91,194,224}
\definecolor{mygreen}{rgb}{0.64, 0.56, 0.88}
\definecolor{myyellow}{rgb}{0.68, 0.6, 0.1}
\definecolor{fancygreen}{rgb}{0.33, 0.68, 0.20}
\definecolor{salmon}{rgb}{0.94, 0.52, 0.49}
\definecolor{tablegreen}{rgb}{0.82, 0.94, 0.75}
\definecolor{tableblue}{rgb}{0.81, 0.90, 0.94}
\definecolor{tablered}{rgb}{0.97, 0.85, 0.85}
\definecolor{tableorange}{rgb}{0.96, 0.85, 0.81}
\definecolor{myorange}{rgb}{1.0, 0.49, 0.0}
\definecolor{tlgreen}{rgb}{0.33, 0.68, 0.20}
\definecolor{darkgreen}{RGB}{0,100,0}
\definecolor{darkred}{RGB}{200, 0, 0}
\definecolor{customyellow}{HTML}{FFFACD}
\definecolor{refinegreen}{RGB}{0, 128, 75}
\definecolor{scoregreen}{RGB}{34, 139, 34}
\definecolor{hidden-blue}{RGB}{194,232,247}
\definecolor{hidden-black}{RGB}{20,68,106}
\definecolor{yes}{HTML}{C6EFCE}
\definecolor{no}{HTML}{FFC7CE}
\definecolor{partial}{HTML}{FFEB9C}
\definecolor{external}{HTML}{D9E1F2}
\definecolor{hdr}{HTML}{F2F2F2}
\definecolor{GRPOrow}{gray}{0.96}
\definecolor{FlowRLrow}{RGB}{225,236,255}
\definecolor{FlowBlue}{RGB}{80,120,210}
\definecolor{GRPOGray}{gray}{0.35}

\hypersetup{
    colorlinks=true, 
    citecolor=uclablue, 
    linkcolor=bigaired,
    urlcolor=darkblue
}

\setlist[itemize]{leftmargin=20pt, noitemsep, topsep=0pt}

\NewDocumentCommand{\kaiyan}{mO{}}{\textcolor{purple}{\textsuperscript{\textit{kaiyan}}\textsf{\textbf{\small[#1]}}}}
\NewDocumentCommand{\yuxin}{mO{}}{\textcolor{cyan}{\textsuperscript{\textit{yuxin}}\textsf{\textbf{\small[#1]}}}}
\NewDocumentCommand{\bx}{mO{}}{\textcolor{green}{\textsuperscript{\textit{bx}}\textsf{\textbf{\small[#1]}}}}
\NewDocumentCommand{\at}{mO{}}{\textcolor{red}{\textsuperscript{\textit{AT}}\textsf{\textbf{\small[#1]}}}}
\NewDocumentCommand{\re}{mO{}}{\textcolor{blue}{\textsuperscript{\textit{RE}}\textsf{\textbf{\small[#1]}}}}
\NewDocumentCommand{\ybsun}{mO{}}{\textcolor{magenta}{\textsuperscript{\textit{youbang}}\textsf{\textbf{\small[#1]}}}}
\NewDocumentCommand{\runze}{mO{}}{\textcolor{orange}{\textsuperscript{\textit{runze}}\textsf{\textbf{\small[#1]}}}}
\NewDocumentCommand{\add}{mO{}}{\textcolor{darkgreen}{\textsuperscript{\textit{Maybe Consider Discuss}}\textsf{\textbf{[#1]}}}}

\newcommand{\cmark}{\textcolor{darkgreen}{\boldmath$\checkmark$}}
\newcommand{\xmark}{\textcolor{darkred}{\boldmath$\times$}}

\newenvironment{itemize*}%
 {\leftmargini=10pt\begin{itemize}%
  \setlength{\itemsep}{0pt}%
  \setlength{\parskip}{0pt}%
  }%
 {\end{itemize}}

\newenvironment{enumerate*}%
 {\begin{enumerate}%
  \setlength{\itemsep}{0pt}%
  \setlength{\parskip}{0pt}}%
 {\end{enumerate}}

\newcommand{\cellstatus}[1]{%
  \begingroup
  \StrTrim{#1}[\statusval]%
  \IfStrEq{\statusval}{Yes}{\cellcolor{yes}\cmark}{}%
  \IfStrEq{\statusval}{No}{\cellcolor{no}\xmark}{}%
  \IfBeginWith{\statusval}{Yes (}{\cellcolor{yes}\cmark~\textit{\statusval\unskip}}{}%
  \IfStrEq{\statusval}{Partial}{\cellcolor{partial}\textbf{Partial}}{}%
  \IfStrEq{\statusval}{External}{\cellcolor{external}\textbf{External}}{}%
  \endgroup
}

\newtcolorbox{myboxi}[1][]{
  breakable,
  title=#1,
  colback=red!5,
  colbacktitle=red!5,
  coltitle=black,
  fonttitle=\bfseries,
  bottomrule=0pt,
  toprule=0pt,
  leftrule=2pt,
  rightrule=2pt,
  titlerule=0pt,
  arc=0pt,
  outer arc=0pt,
  colframe=red,
}

\newtcolorbox{myboxnote}[1][]{
  breakable,
  title=#1,
  colback=orange!0,
  colbacktitle=orange!0,
  coltitle=black,
  fonttitle=\bfseries,
  bottomrule=0pt,
  toprule=0pt,
  leftrule=2pt,
  rightrule=2pt,
  titlerule=0pt,
  arc=0pt,
  outer arc=0pt,
  colframe=orange,
}

\newtcolorbox{myboxii}[1][]{
  breakable,
  freelance,
  title=#1,
  colback=white,
  colbacktitle=white,
  coltitle=black,
  fonttitle=\bfseries,
  bottomrule=0pt,
  boxrule=0pt,
  colframe=white,
  overlay unbroken and first={
  \draw[red!75!black,line width=3pt]
    ([xshift=5pt]frame.north west) -- 
    (frame.north west) -- 
    (frame.south west);
  \draw[red!75!black,line width=3pt]
    ([xshift=-5pt]frame.north east) -- 
    (frame.north east) -- 
    (frame.south east);
  },
  overlay unbroken app={
  \draw[red!75!black,line width=3pt,line cap=rect]
    (frame.south west) -- 
    ([xshift=5pt]frame.south west);
  \draw[red!75!black,line width=3pt,line cap=rect]
    (frame.south east) -- 
    ([xshift=-5pt]frame.south east);
  },
  overlay middle and last={
  \draw[red!75!black,line width=3pt]
    (frame.north west) -- 
    (frame.south west);
  \draw[red!75!black,line width=3pt]
    (frame.north east) -- 
    (frame.south east);
  },
  overlay last app={
  \draw[red!75!black,line width=3pt,line cap=rect]
    (frame.south west) --
    ([xshift=5pt]frame.south west);
  \draw[red!75!black,line width=3pt,line cap=rect]
    (frame.south east) --
    ([xshift=-5pt]frame.south east);
  },
}

\tcbset{
  takeawaysbox/.style={
    title=Takeaways,
    colback=lightblue!80,
    colframe=black,
    fonttitle=\bfseries\small,
    coltitle=white,
    colbacktitle=black,
    enhanced,
    attach boxed title to top left={xshift=2.5mm,yshift=-2.5mm},
    boxed title style={rounded corners, size=small, colframe=black, colback=black},
    width=\linewidth,
    arc=3.5mm
  }
}

\mdfdefinestyle{mystyle}{%
  rightline=true,
  innerleftmargin=10,
  innerrightmargin=10,
  outerlinewidth=3pt,
  topline=false,
  rightline=true,
  bottomline=false,
  skipabove=\topsep,
  skipbelow=\topsep
}

\tikzset{%
    every node/.style={font=\tiny},
    parent/.style =          {align=center,text width=2cm,rounded corners=3pt, line width=0.3mm, fill=gray!10,draw=gray!80},
    child/.style =           {align=center,text width=2.0cm,rounded corners=3pt, fill=blue!10,draw=blue!80,line width=0.3mm},
    grandchild/.style =      {align=center,text width=2cm,rounded corners=3pt},
    greatgrandchild/.style = {align=center,text width=1.5cm,rounded corners=3pt},
    greatgrandchild2/.style = {align=center,text width=1.5cm,rounded corners=3pt},    
    referenceblock/.style =  {align=center,text width=1.5cm,rounded corners=2pt},
    pretrain/.style =           {align=center,text width=2.0cm,rounded corners=3pt, fill=blue!10,draw=blue!80,line width=0.3mm},   
    pretrain_work/.style =           {align=center, text width=8.5cm,rounded corners=3pt, fill=blue!10,draw=blue!0,line width=0.3mm},  
    template/.style =           {align=center,text width=2.0cm,rounded corners=3pt, fill=red!10,draw=red!80,line width=0.3mm},   
    template_work/.style =           {align=center,text width=8.5cm,rounded corners=3pt, fill=red!10,draw=red!0,line width=0.3mm},    
    answer/.style =           {align=center,text width=2.0cm,rounded corners=3pt, fill= cyan!10,draw= cyan!80,line width=0.3mm},   
    answer_work/.style =           {align=center,text width=8.5cm,rounded corners=3pt, fill= cyan!10,draw= cyan!0,line width=0.3mm},      
    multiple/.style =           {align=center,text width=2.0cm,rounded corners=3pt, fill= orange!10,draw= orange!80,line width=0.3mm},   
    multiple_work/.style =           {align=center,text width=8.5cm,rounded corners=3pt, fill= orange!10,draw= orange!0,line width=0.3mm},        
    tuning/.style =           {align=center,text width=2.0cm,rounded corners=3pt, fill= magenta!10,draw= magenta!80,line width=0.3mm},   
    tuning_work/.style =           {align=center,text width=8.5cm,rounded corners=3pt, fill= magenta!10,draw= magenta!0,line width=0.3mm},          
}

\newcommand{\lstbg}[3][0pt]{{\fboxsep#1\colorbox{#2}{\strut #3}}}

\lstdefinelanguage{diff}{
  basicstyle=\ttfamily\small,
  morecomment=[f][\lstbg{red!20}]-,
  morecomment=[f][\lstbg{green!20}]+,
}

\lstdefinelanguage{diffpython}{
  language=diff,
  morekeywords={def, if, else, for, while, return, import, from, as, class, with, try, except, finally, raise, lambda, and, or, not, in, is, None, True, False},
  morecomment=[l]{\#},
  morestring=[b]",
  morestring=[b]',
}

\setheadertext{LUMIA Lab}

\correspondingemail{\emailicon \{whyisverysmart, lin.zhouhan\}@gmail.com \quad $^\ddagger$ Corresponding Author. }

\githublink{https://github.com/whyisverysmart/Fourier-Compressor}

\renewcommand{\today}{}

\title{\fontsize{20.5}{23}\selectfont
Fourier Compressor: Frequency-Domain Visual Token Compression for Vision-Language Models}
\setheadertitle{Fourier Compressor}

\author{%
  \Authfont Huanyu Wang$^{1,4}$, Jushi Kai$^{1,4}$, Haoli Bai$^{3}$, Lu Hou$^{3}$, Bo Jiang$^{4}$, Ziwei He$^{2,\ddagger}$, Zhouhan Lin$^{1,2,\ddagger}$\\
  $^{1}$ LUMIA Lab, School of Artificial Intelligence, Shanghai Jiao Tong University\\
  $^{2}$ Shanghai Innovation Institute\\
  $^{3}$ Noah’s Ark Lab, Huawei Technologies Ltd.\\
  $^{4}$ School of Computer Science, Shanghai Jiao Tong University\\
}

\begin{document}

\begin{abstract}

Vision-Language Models (VLMs) incur substantial computational overhead and inference latency due to the large number of vision tokens introduced by high-resolution image and video inputs. Existing parameter-free token compression methods typically rely on token selection or merging, yet they risk discarding substantial visual information or distorting the original representation distribution, resulting in pronounced performance degradation at high compression ratios. In response, we aim to explore a more effective and efficient visual token compression strategy, with a promising direction in the frequency domain. Motivated by the success of frequency-domain transforms in image compression (\eg, JPEG), we systematically analyze the frequency redundancy in visual representations and uncover a non-uniform distribution of semantic information across frequency bands. Building upon this, we introduce \textbf{Fourier Compressor}, an effective, parameter-free, and highly generalizable module that removes redundancy from visual representations within the frequency domain. Implemented via FFT with $\mathcal{O}(n^2 \log n)$ complexity and no additional parameters, Fourier Compressor introduces negligible computational overhead while preserving semantic fidelity. Extensive experiments on image-based benchmarks demonstrate that our method achieves a favorable performance-efficiency trade-off, retaining over 96\% of the original accuracy while reducing inference FLOPs by up to 83.8\% and boosting generation speed by 31.2\%. It consistently outperforms existing parameter-free methods and even surpasses some parameterized approaches. Importantly, Fourier Compressor generalizes consistently across both LLaVA and Qwen-VL architectures, and further extends to video understanding tasks, highlighting its practical applicability for efficient VLMs.

\end{abstract}

\maketitle


\section{Introduction}

Vision-Language Models (VLMs) extend Large Language Models (LLMs) with visual understanding capabilities by integrating a vision encoder via a “glue layer” (\eg, Multi-Layer Perceptron or Q-Former~\citep{BLIP2}) to jointly model visual and textual information. Nevertheless, the generation of a large number of vision tokens imposes substantial computational demands, as the backbone LLM must process an extremely long context, resulting in significant overhead and high inference latency. For instance, LLaVA-v1.5~\citep{LLaVA-v1.5} produces 576 visual tokens for a single $336 \times 336$ image, while Qwen-VL series~\citep{qwen2vl, qwen25vl} generates one token per $28 \times 28$ pixels, producing several thousand tokens for high-resolution images that often constitute the majority of the input context, far exceeding the number of text tokens. This challenge is further amplified when processing multiple images and video inputs.

Fortunately, visual representations in VLMs exhibit considerable redundancy, motivating the development of vision token compression. Existing parameter-free methods generally fall into two paradigms. Approaches such as FastV~\citep{fastv}, ATP-LLaVA~\citep{ATP-LLaVA}, and SparseVLM~\citep{sparsevlm} rely on heuristic token selection, which risks discarding informative visual content. Others, including VisionZip~\citep{visionzip} and LLaVA-PruMerge~\citep{LLaVA-PruMerge}, adopt similarity-based token merging, inevitably perturbing the underlying visual representation distribution. While being computationally efficient, both paradigms suffer pronounced performance degradation under high compression ratios. Parameterized alternatives (\eg, VisToG~\citep{VisToG}, QueCC~\citep{QueCC}, MQT-LLaVA~\citep{MQT-LLaVA}) mitigate these issues with learnable queries, but incur additional compression-induced computational overhead. Extreme designs like LLaVA-mini~\citep{llava-mini} introduce extra transformer blocks, resulting in substantial cost and reduced architectural flexibility.

To address these limitations, we aim to develop a more effective, efficient, and generalizable vision token compression strategy. One promising direction lies in the \textbf{frequency domain}. Frequency-domain transforms have long been exploited in traditional image compression, where they reveal substantial redundancy and enable highly efficient and generalizable compression methods (\eg, JPEG~\citep{jpeg}). Inspired by this, we systematically investigate whether similar frequency redundancy exists in visual representations within VLMs. Our analysis in Section~\ref{subsec:analyses} confirms the presence of such redundancy, uncovering a non-uniform distribution of semantic information across frequency bands. This property can be directly exploited to design parameter-free token compression strategies that are broadly applicable across different VLM architectures.

To this end, we introduce \textbf{Fourier Compressor}, an effective, parameter-free, and highly generalizable module for compressing vision tokens in VLMs. By exploiting frequency-domain redundancy in visual representations, our method achieves a favorable trade-off, significantly reducing token counts at high efficiency while preserving generation performance. Fourier Compressor introduces no additional parameters, resulting in minimal compression-induced computational overhead, and generalizes well across multiple VLM architectures, offering an efficient and broadly applicable solution for vision token compression.

The rest of this paper is organized as follows. Section~\ref{sec:related} reviews related work. Section~\ref{sec:preliminaries} presents preliminaries on the Discrete Cosine Transform and the analyses of frequency components in visual representations. Section~\ref{sec:method} describes the detailed design  and theoretical time complexity analysis of our method. Section~\ref{sec:experiments} presents performances on image-based benchmarks, evaluates empirical efficiency, and demonstrates applicability to video tasks. Finally, Section~\ref{sec:conclusion} concludes the paper.

\section{Related Work}
\label{sec:related}

\subsection{Visual Token Compression in VLMs}

Vision-Language Models (VLMs) have achieved remarkable progress in image and video understanding by integrating LLMs with visual encoders. Despite strong performance from models like LLaVA series~\citep{LLaVA-v1, LLaVA-v1.5} and Qwen-VL series~\citep{qwen2vl, qwen25vl}, the high volume of vision tokens remains a major bottleneck for efficient inference and practical deployment.

To address the challenge of long contexts in VLMs, previous research has primarily focused on reducing the number of vision tokens. One common approach involves selecting the most relevant vision tokens or merging less important ones. For example, ATP-LLaVA~\citep{ATP-LLaVA} computes an importance score for each vision token and dynamically determines a pruning threshold to remove redundant tokens within the backbone LLM, while LLaVA-PruMerge~\citep{LLaVA-PruMerge} clusters and averages vision tokens according to similarity between the class token and spatial tokens. Other methods leverage learnable parameters to extract visual features. MQT-LLaVA~\citep{MQT-LLaVA} employs a Matryoshka Query Transformer, whereas QueCC~\citep{QueCC} introduces a query-based convolutional cross-attention module that enables text embeddings to query vision tokens. Furthermore, recent research has also explored transferring visual information to language tokens, such as LLaVA-mini~\citep{llava-mini}, which employs a prefusion module that allows text tokens to integrate relevant visual information in advance. However, these approaches have yet to find an optimal balance between the additional computational cost of compression and performance degradation. This challenge motivates us to further investigate cost-efficient token compression methods.

\subsection{Frequency-based Compression}

Frequency domain techniques have long played an important role in signal processing and data compression. In computer vision, frequency transformations are foundational to standard image compression algorithms, such as JPEG~\citep{jpeg}. Previous study~\citep{FreqDomain} has also shown that convolutional neural networks (CNNs) exhibit a strong sensitivity to low-frequency channels, revealing an inherent frequency bias in visual feature extraction. Beyond the vision domain, frequency-based approaches have also been widely explored in natural language processing. For instance, FNet~\citep{fnet} replaces the self-attention mechanism in Transformer-like encoders with frequency transformation, achieving comparable performance with significantly reduced computational costs. Fourier-Transformer~\citep{fourier-transformer} applies frequency-domain truncation to downsample hidden states in Transformer models for improved efficiency. More recently, FreqKV~\citep{freqkv} introduces a frequency-based key-value compression technique that effectively extends the context window of LLMs. When it comes to VLMs, DocPedia~\citep{docpedia} leverages DCT coefficients extracted directly from RGB images to perform vision encoding, enabling higher input resolutions for document understanding tasks. However, the potential of applying frequency-domain techniques to visual-token-level representations remains largely unexplored.

\section{Preliminaries}
\label{sec:preliminaries}

In this section, we introduce the preliminaries of our method, including the formulation of the Discrete Cosine Transform (Section~\ref{subsec:dct}), and the analyses of frequency components in visual representations that reveal semantic redundancy within the frequency domain (Section~\ref{subsec:analyses}).

\subsection{Discrete Cosine Transform}
\label{subsec:dct}

The Discrete Cosine Transform (DCT) is a linear invertible function $\mathbf{\Psi}\colon \mathbb{R}^{n} \rightarrow \mathbb{R}^{n}$ that converts a sequence of discrete real numbers from the spatial domain to the frequency domain. It exhibits strong energy compaction properties, as most signal information (\eg, audio, image) tends to concentrate in the low-frequency components after transformation. Among the various variants of DCT, we conduct the most widely used type-II DCT.

Formally, for a real-valued sequence $\langle x_{i}\rangle=\{x_{0}, x_{1}, \dots, x_{N-1}\}$, the DCT transformation is given by:
\begin{equation}
    f_{m} = \alpha_{m}\sum^{N-1}_{i=0} x_{i}\cdot \phi_{N}(m, i),
\end{equation}
where $m\in \{0, 1, \dots, N-1\}$. The basis function $\phi_{\cdot}(\cdot, \cdot)$ and the normalization factor $\alpha_{m}$ are defined as:
\begin{equation}
    \begin{aligned}
        & \phi_{N}(x, y) = \cos\left[\frac{\pi}{N}x\left(y+\frac{1}{2}\right)\right] \\
        & \alpha_{m} = 
            \begin{cases}
                \sqrt{\frac{1}{N}} & \textit{if m = 0,}\\
                \sqrt{\frac{2}{N}} & \textit{otherwise.}
            \end{cases}
    \end{aligned}
    \label{eq:DCT}
\end{equation}

The above transformation expresses $\langle x_{i}\rangle$ as a sum of orthogonal cosine functions at different frequencies, where the coefficients represent the contribution of each frequency component.

Conversely, given the frequency representation $\langle f_{m}\rangle=\{f_{0}, f_{1}, \dots, f_{N-1}\}$, the original sequence can be recovered by the inverse Discrete Cosine Transform (iDCT):
\begin{equation}
    x_{i} = \sum^{N-1}_{k=0} \alpha_{k}\cdot f_{k}\cdot \phi_{N}(k, i)
\end{equation}

\subsection{Analyses of Frequency Components}
\label{subsec:analyses}

To systematically investigate the potential frequency redundancy of visual representations and examine the semantic information captured by different frequency bands, we conduct three complementary analyses focusing on: (i) \emph{energy distribution}, (ii) \emph{image perturbation}, and (iii) \emph{semantic-continuous morphing}.

\begin{figure}[t]
    \centering
    \begin{subfigure}{0.15\linewidth}
        \centering
        \includegraphics[width=\linewidth]{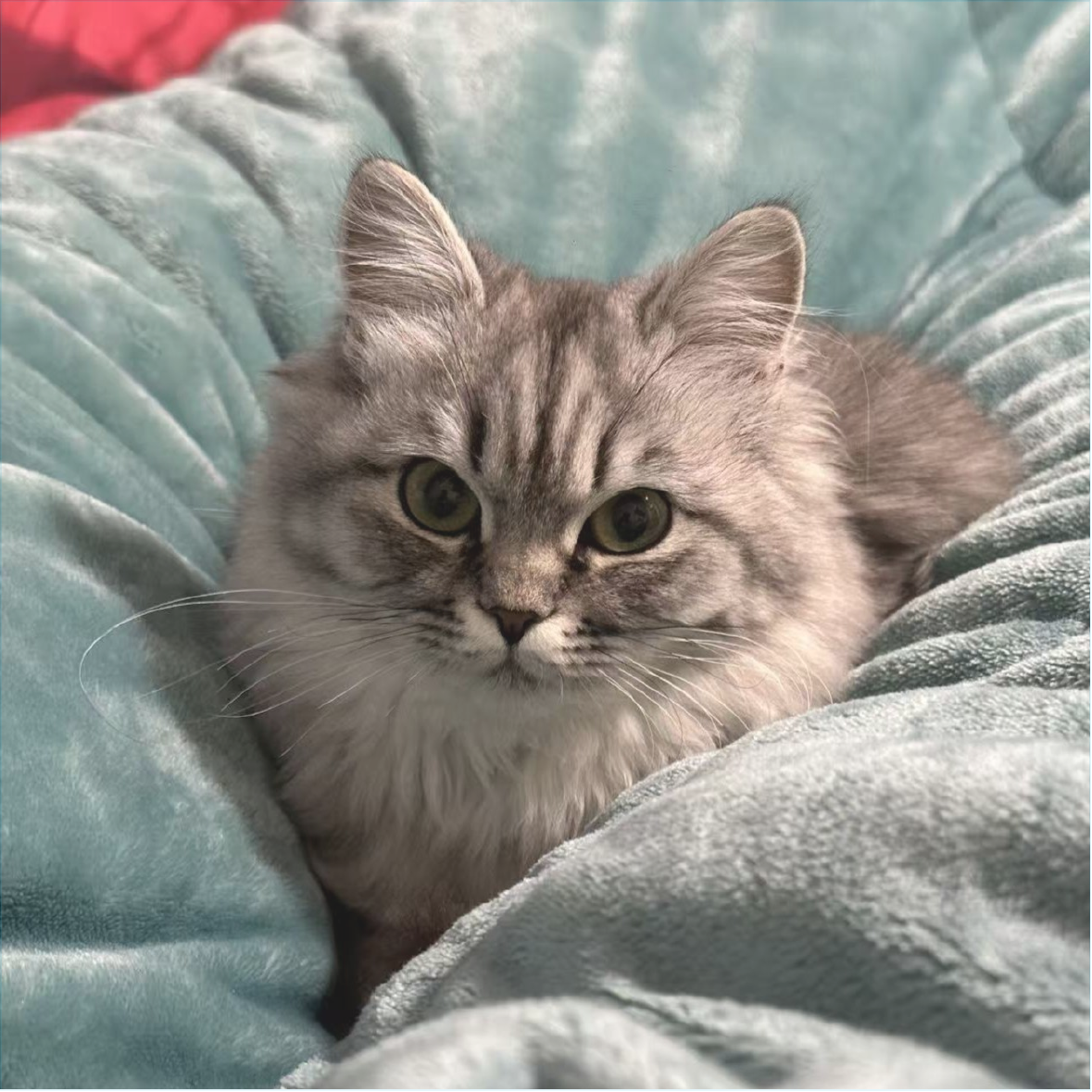}\\
        \vspace{0.4em}
        \includegraphics[width=\linewidth]{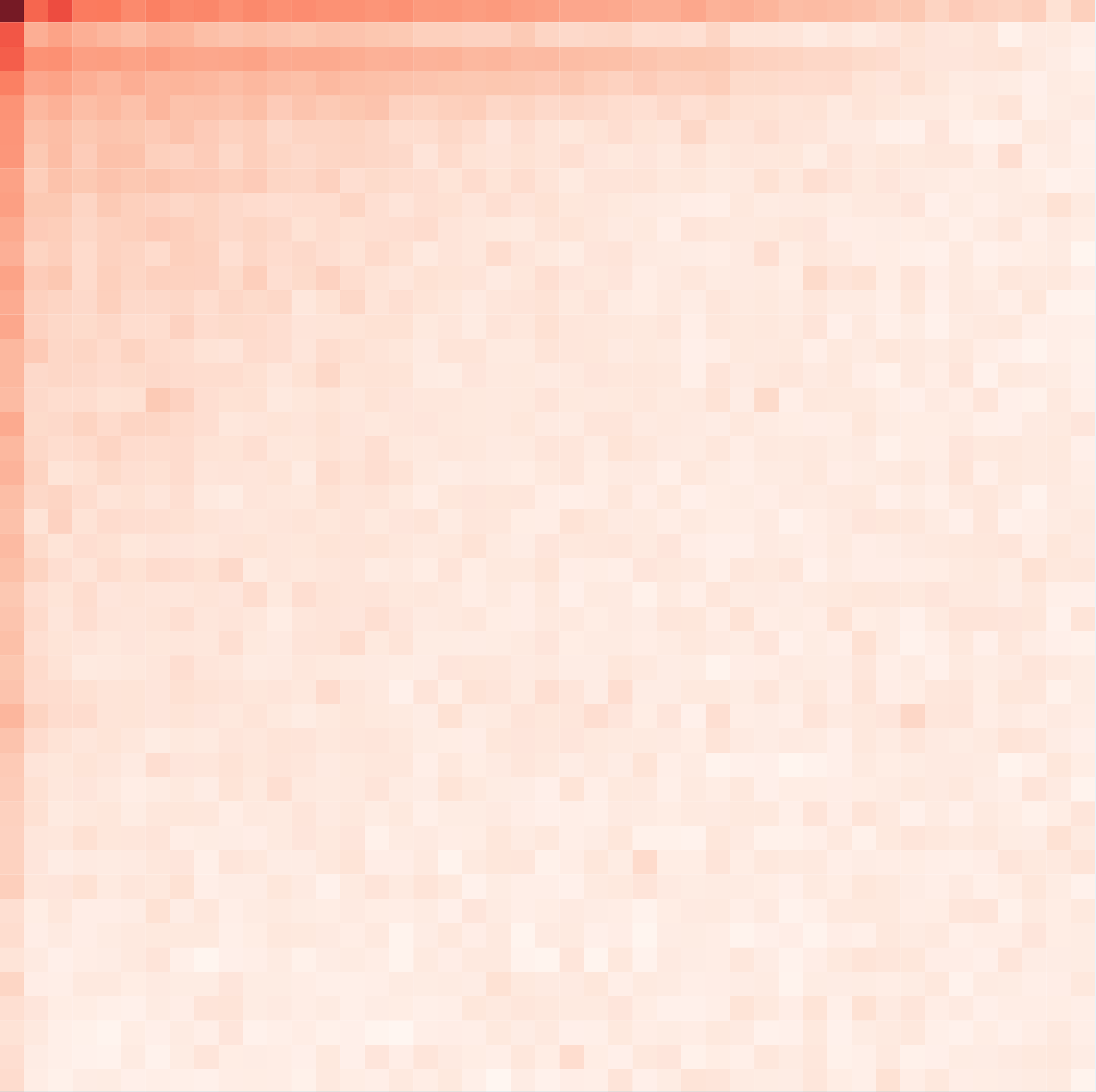}
        \caption{Cat}
        \label{fig:cat}
    \end{subfigure}
    \hfill
    \begin{subfigure}{0.15\linewidth}
        \centering
        \includegraphics[width=\linewidth]{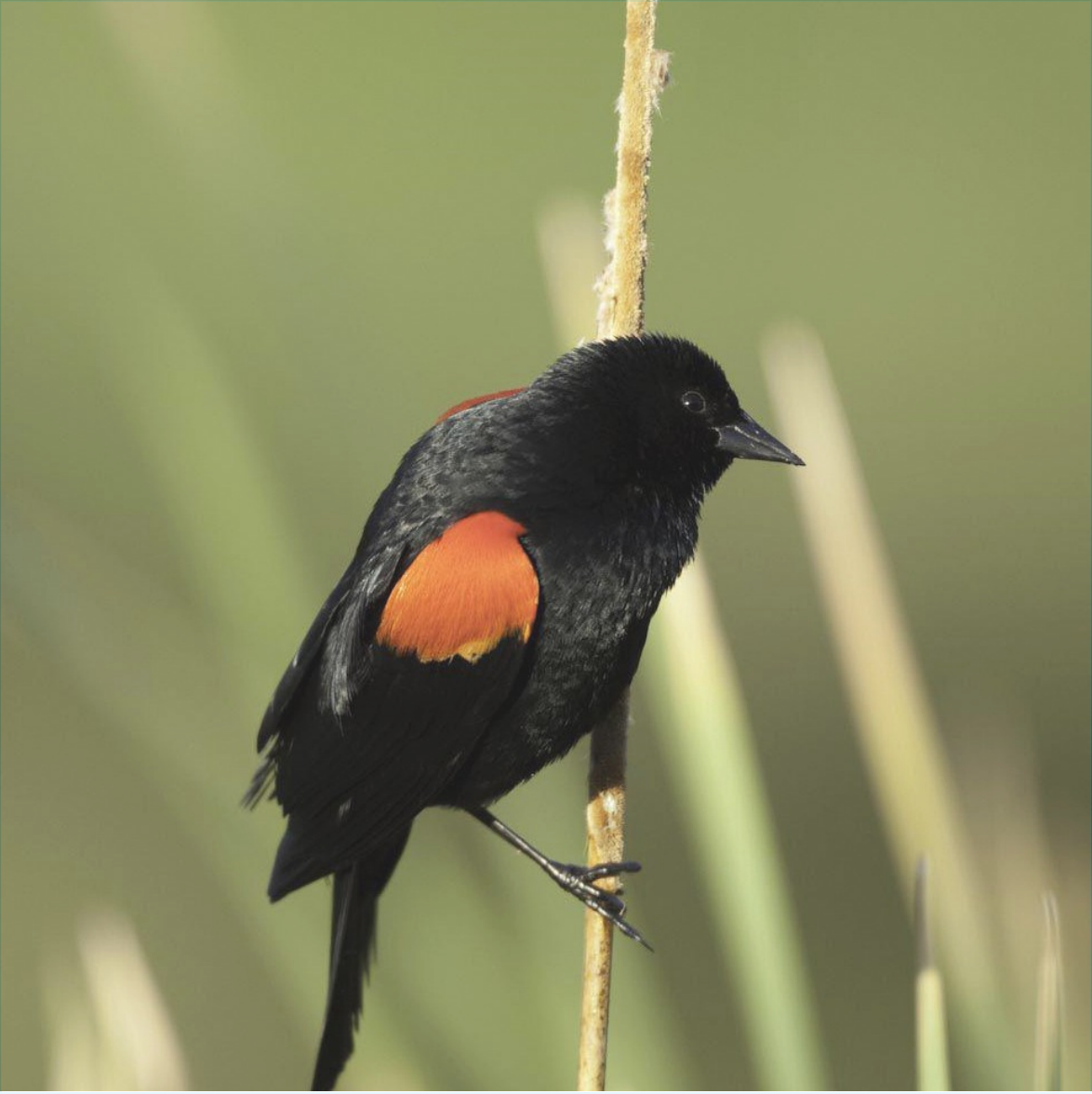}\\
        \vspace{0.4em}
        \includegraphics[width=\linewidth]{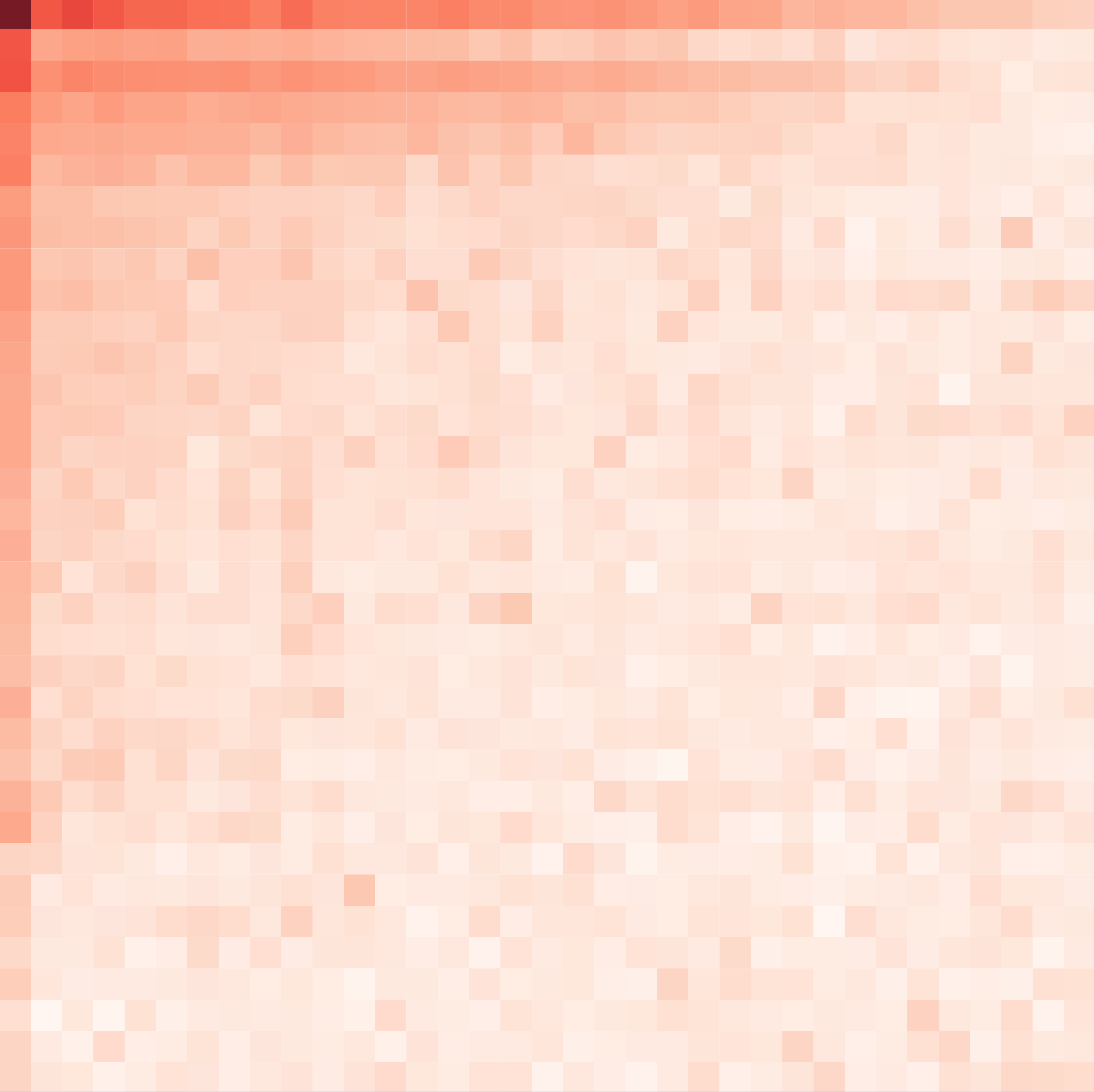}
        \caption{Bird}
        \label{fig:bird}
    \end{subfigure}
    \hfill
    \begin{subfigure}{0.15\linewidth}
        \centering
        \includegraphics[width=\linewidth]{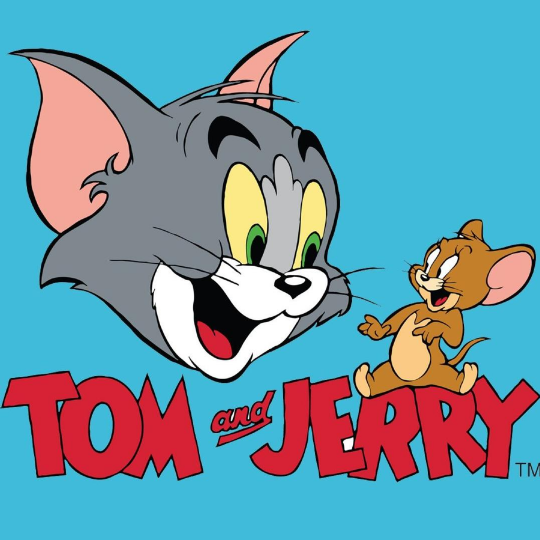}\\
        \vspace{0.4em}
        \includegraphics[width=\linewidth]{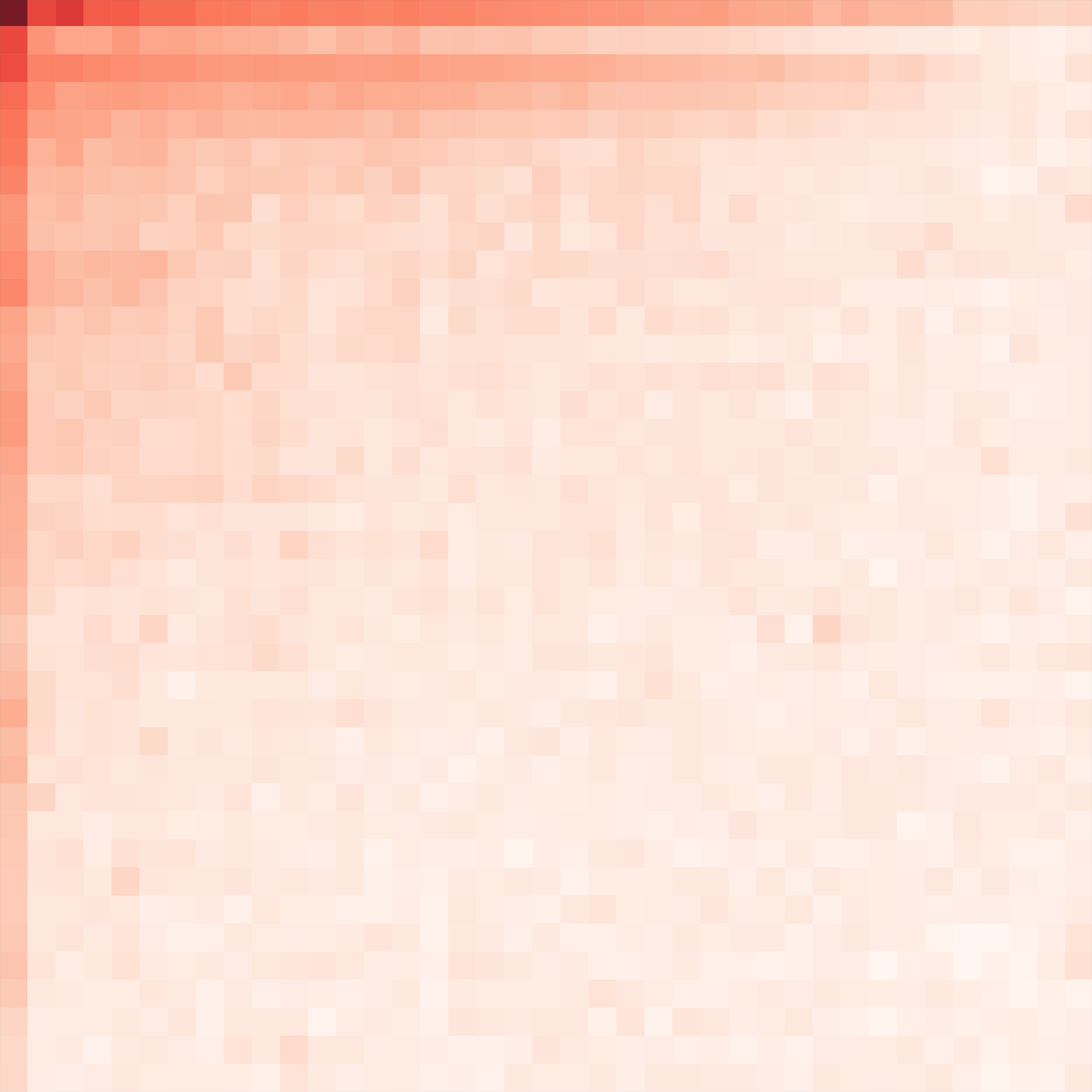}
        \caption{Anime}
        \label{fig:anime}
    \end{subfigure}
    \hfill
    \begin{subfigure}{0.15\linewidth}
        \centering
        \includegraphics[width=\linewidth]{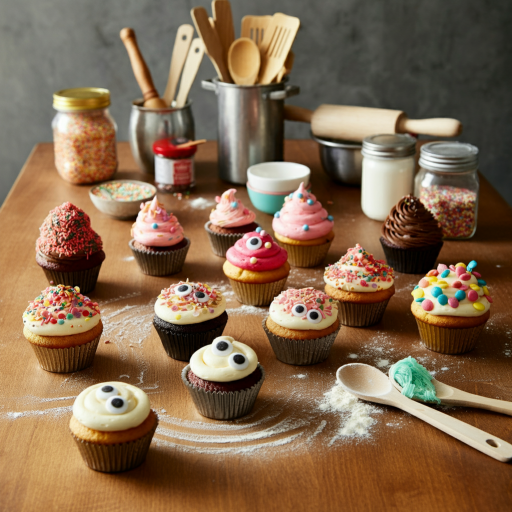}\\
        \vspace{0.4em}
        \includegraphics[width=\linewidth]{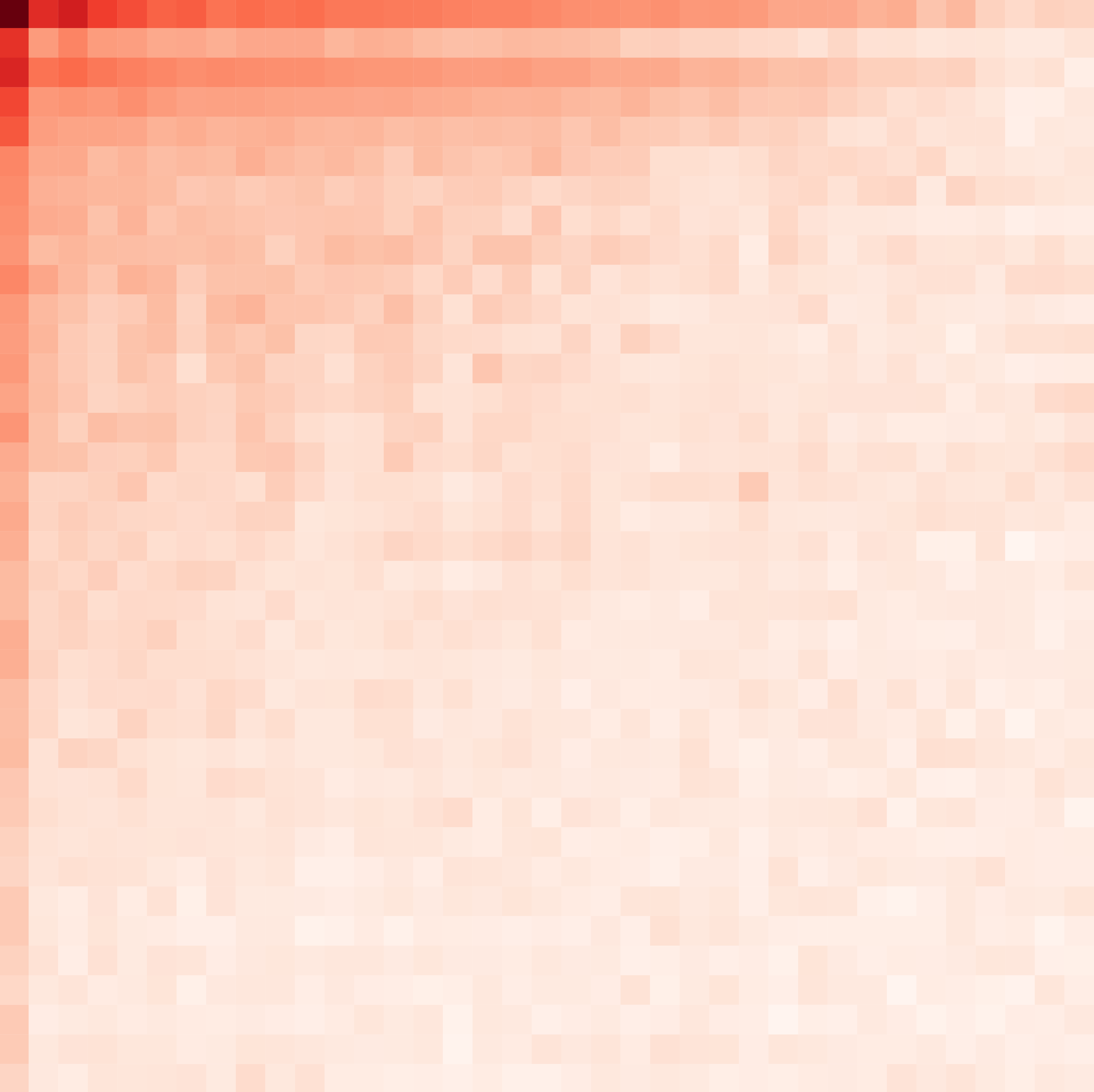}
        \caption{Cakes}
        \label{fig:cakes}
    \end{subfigure}
    \hfill
    \begin{subfigure}{0.15\linewidth}
        \centering
        \includegraphics[width=\linewidth]{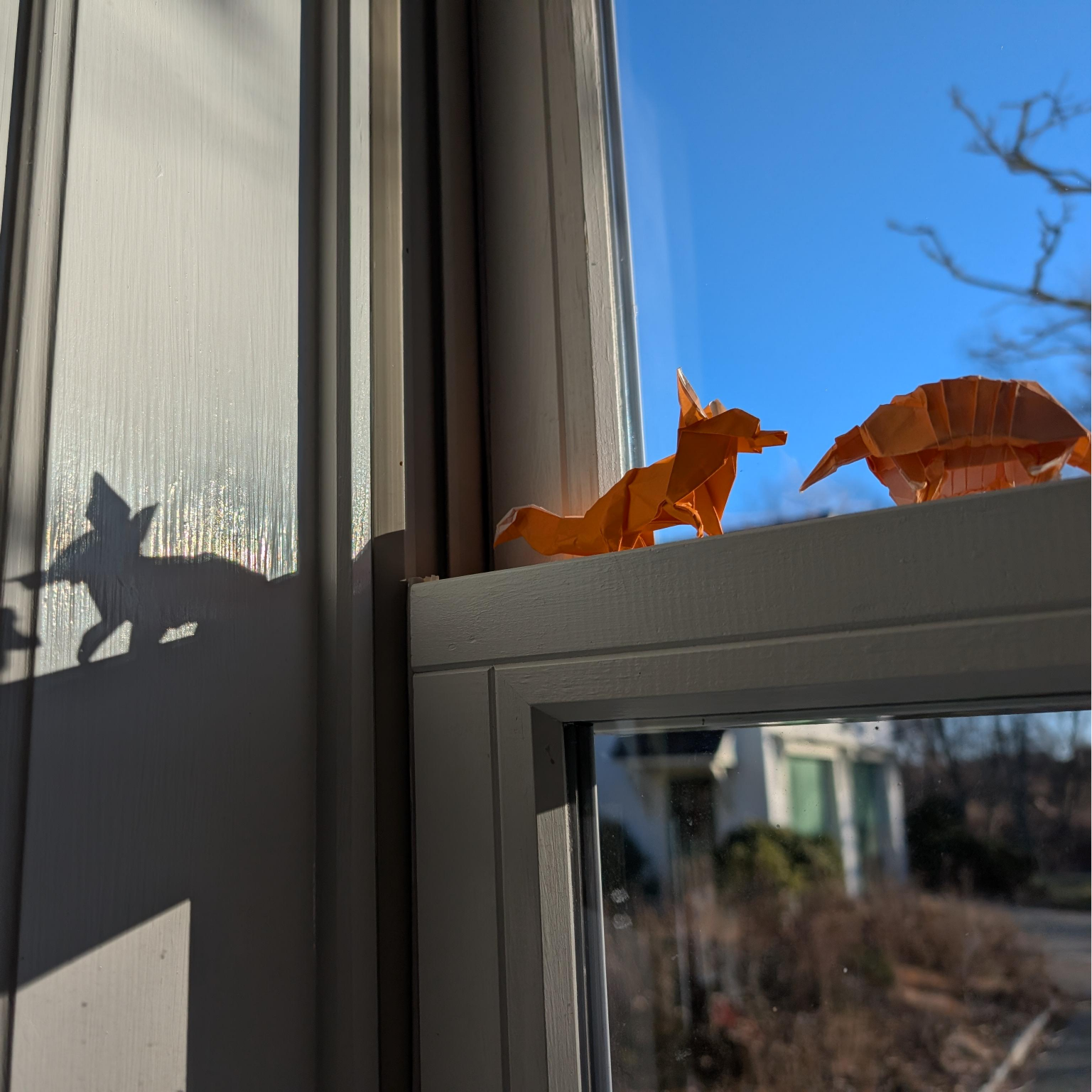}\\
        \vspace{0.4em}
        \includegraphics[width=\linewidth]{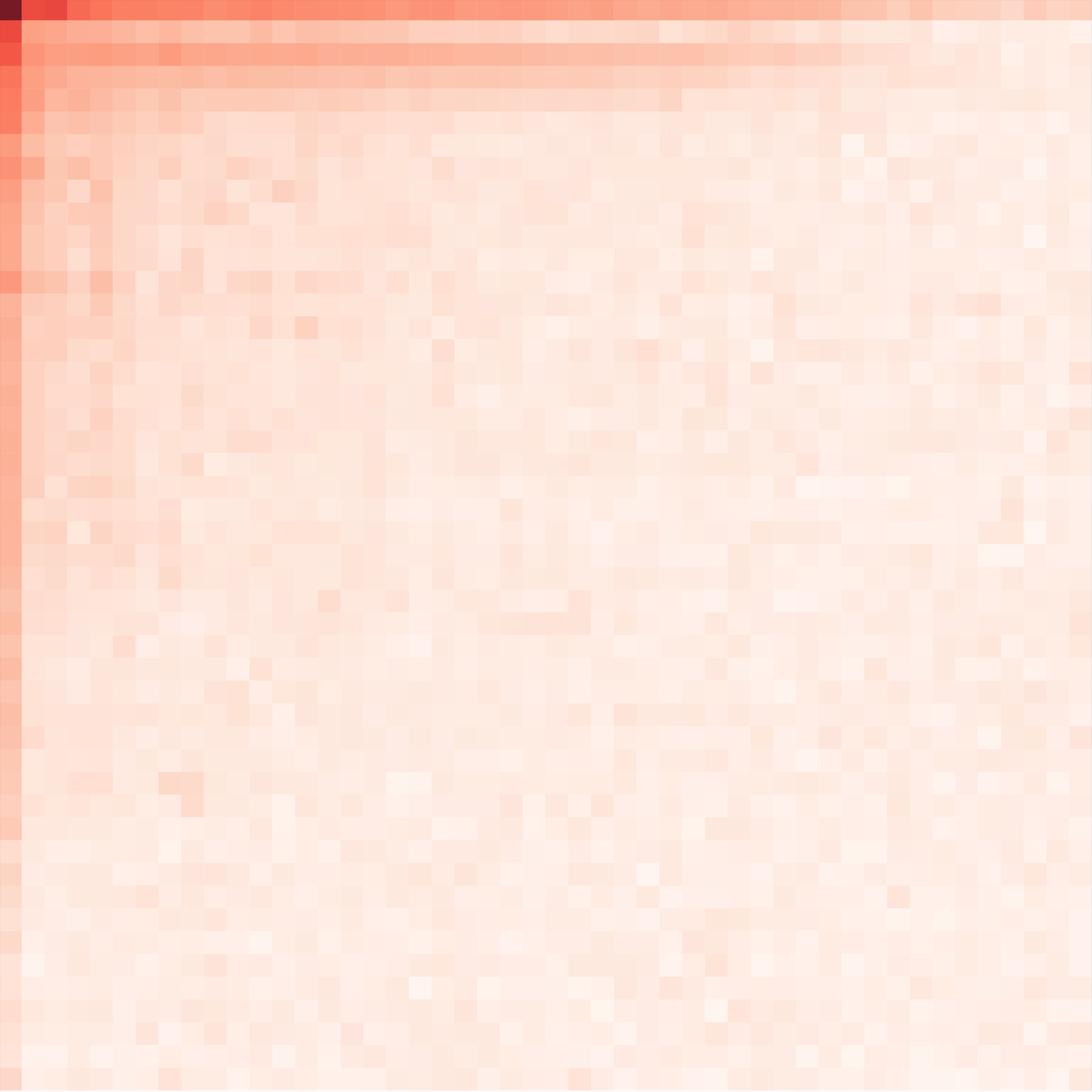}
        \caption{Origami}
        \label{fig:origami}
    \end{subfigure}
    \hfill
    \begin{minipage}[t]{0.01\linewidth}
        \centering
        \begin{tikzpicture}
            \draw[densely dashed] (0,0) -- (0,5.75cm);
        \end{tikzpicture}
    \end{minipage}
    \hfill
    \begin{subfigure}{0.15\linewidth}
        \centering
        \includegraphics[width=\linewidth]{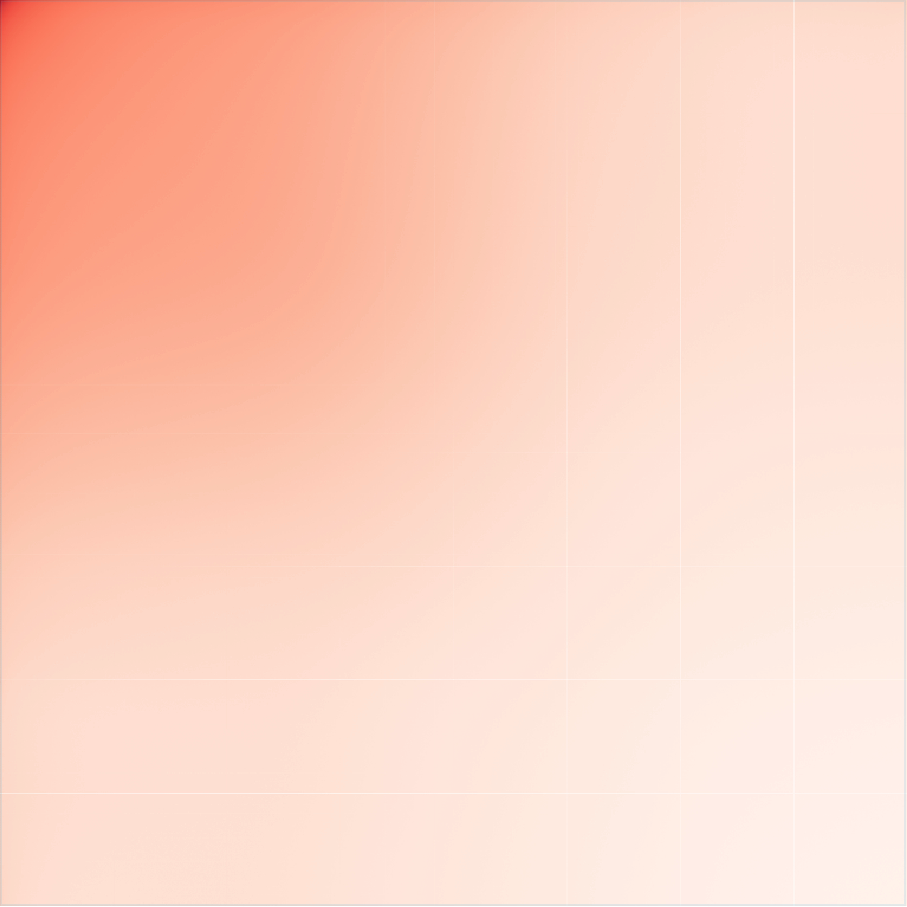}\\
        \vspace{0.4em}
        \includegraphics[width=\linewidth]{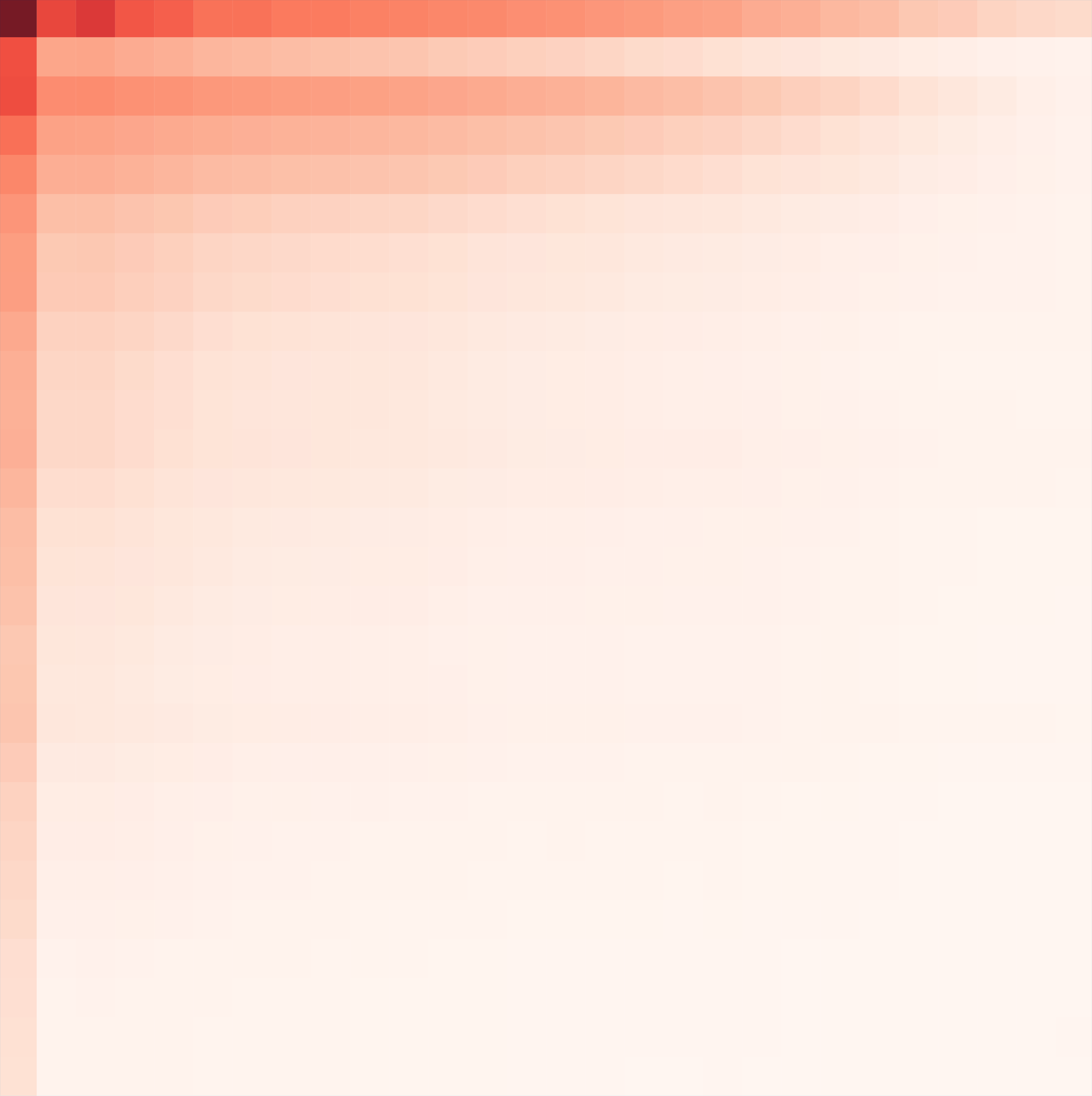}
        \caption{Avg.}
        \label{fig:avg}
    \end{subfigure}
    \caption{Heatmap visualization of the frequency spectra of visual representations computed by the Qwen2-VL vision encoder. We consider only the magnitude, averaging the absolute values across all hidden dimensions for each frequency component, which is then plotted on a logarithmic scale. In (f), the upper and lower heatmaps show the averaged frequency spectra of raw images and their corresponding visual representations from 10,000 sampled images in GQA, respectively.}
    \label{fig:heatmap}
\end{figure}

\textbf{Energy distribution.} For multiple images, we apply the two-dimensional Discrete Cosine Transform (DCT) to the visual representations produced by the Qwen2-VL~\citep{qwen2vl} vision encoder, and compute the energy of each frequency component by averaging the absolute values across all hidden dimensions. As shown in Figure~\ref{fig:heatmap}, the energy of these visual representations exhibits \textbf{a strong concentration in low-frequency components} across all hidden dimensions. The prominence of this pattern varies across images of different types, suggesting that the frequency-domain energy distribution inherently encodes structural and semantic information. Importantly, as illustrated in Figure~\ref{fig:avg}, the averaged frequency pattern of visual representations closely resembles that of the corresponding raw images, which motivates our exploration of frequency-domain compression, analogous to JPEG~\citep{jpeg}, for designing more effective and efficient vision token compression strategies.

\begin{wrapfigure}[16]{r}{0.45\linewidth}
    \centering
    \vspace{-1em}
    \includegraphics[width=\linewidth]{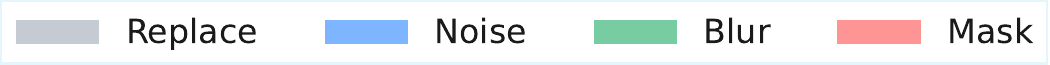}\\
    \begin{subfigure}{\linewidth}
        \centering
        \includegraphics[width=0.85\linewidth]{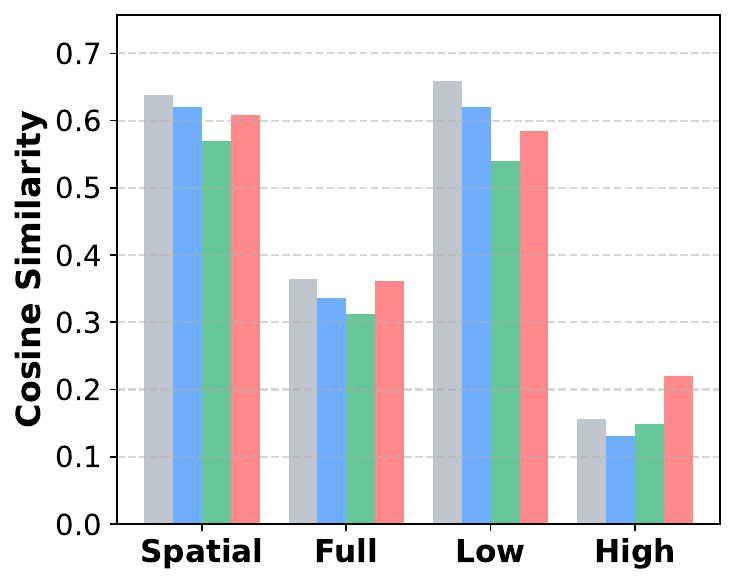}
    \end{subfigure}
    \caption{Cosine similarity of perturbed versus original visual representations for four types of perturbation.}
    \label{fig:perturbation}
\end{wrapfigure}

\textbf{Image Perturbation.} For a fixed input image, we apply four types of perturbation: (a) \emph{Replace}, where a fraction of pixels are replaced with uniform pixel values; (b) \emph{Noise}, adding zero-mean Gaussian noise; (c) \emph{Blur}, applying Gaussian blurring; and (d) \emph{Mask}, randomly masking contiguous square regions up to a target area. We extract the corresponding visual representations using the Qwen2-VL~\citep{qwen2vl} encoder, which are then transformed into the frequency domain via the two-dimensional Discrete Cosine Transform (DCT). We compute the cosine similarity between the perturbed and original images for: spatial vision tokens, full frequency components, and the low- (lowest 25\%) and high- (highest 25\%) frequency components. As illustrated in Figure~\ref{fig:perturbation}, \textbf{low-frequency components consistently demonstrate higher robustness} across all perturbation types, whereas high-frequency components exhibit pronounced sensitivity, suggesting that low-frequency bands predominantly encode global image structures rather than local appearance details.

\begin{wrapfigure}[18]{r}{0.45\linewidth}
    \centering
    \vspace{-1em}
    \includegraphics[width=0.8\linewidth]{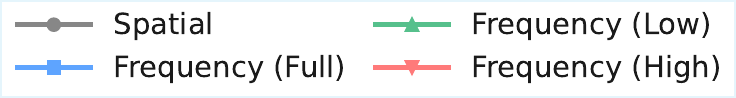}\\
    \includegraphics[width=0.9\linewidth]{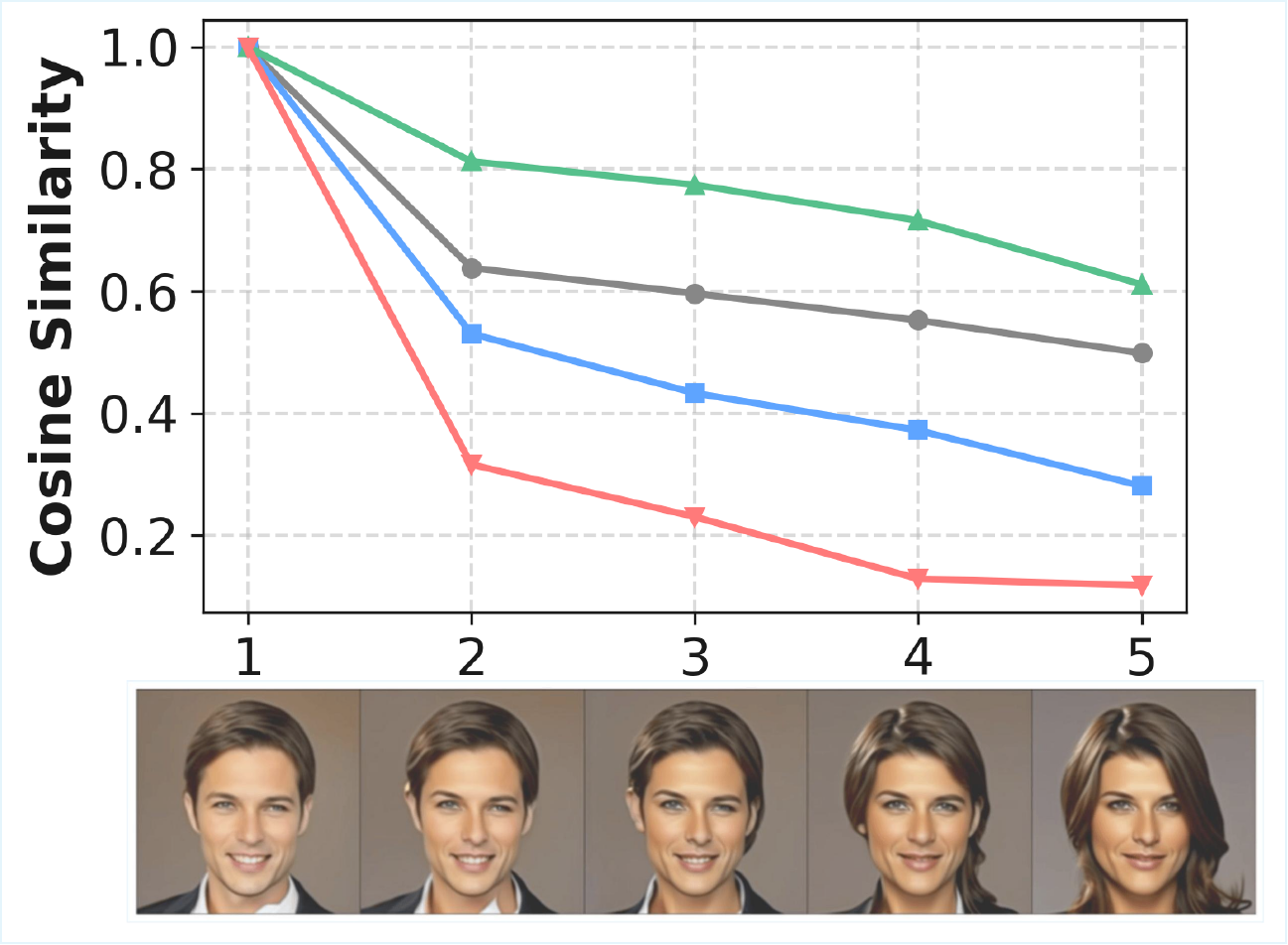}
    \caption{Cosine similarity of visual representations of morphing frames relative to the leftmost image, for spatial tokens, full-, low- (lowest 25\%) and high- (highest 25\%) frequency components.}
    \label{fig:morph}
\end{wrapfigure}

\textbf{Semantic Morphing.} We employ a morphing sequence from FreeMorph~\citep{freemorph}, which exhibits smooth and continuous semantic transitions. Following the same protocol, we compute the cosine similarity between the visual representations of each morphing frame and that of the first (leftmost) image. As shown in~\ref{fig:morph}, \textbf{low-frequency components exhibit substantially greater stability under smooth semantic transitions with consistent global structure} (\eg, frontal pose and facial topology), whereas high-frequency components undergo markedly larger variations, further highlighting their distinct roles in semantic encoding.

\textbf{Conclusion.} These results reveal a clear functional separation across frequency bands and uncover substantial redundancy in the frequency domain of visual representations: Low-frequency components predominantly capture global, coarse-grained semantic attributes, exhibiting strong robustness to perturbations and smooth transitions under semantic changes, and consequently dominate the overall energy distribution; In contrast, high-frequency components are more sensitive to noise and primarily encode fine-grained details. This observation indicates substantial semantic redundancy in visual representations, providing strong empirical motivation for our frequency-domain token compression strategy, which selectively preserves low-frequency information to achieve efficient token reduction while maintaining semantic fidelity.

\section{Method}
\label{sec:method}

In this section, we introduce \textbf{Fourier Compressor}, a parameter-free and highly generalizable module for visual token compression. Building upon a standard VLM architecture, our framework preserves the vision encoder, projector, and backbone LLM, and inserts the Fourier Compressor after the vision encoder to substantially reduce the number of visual tokens. Below we detail the overall architecture (Section~\ref{subsec:architecture}), the design of the Fourier Compressor (Section~\ref{subsec:module}), and the theoretical time complexity analysis (Section~\ref{subsec:complexity}).

\subsection{General Architecture}
\label{subsec:architecture}

\begin{figure}[t]
    \centering
    \includegraphics[width=\linewidth]{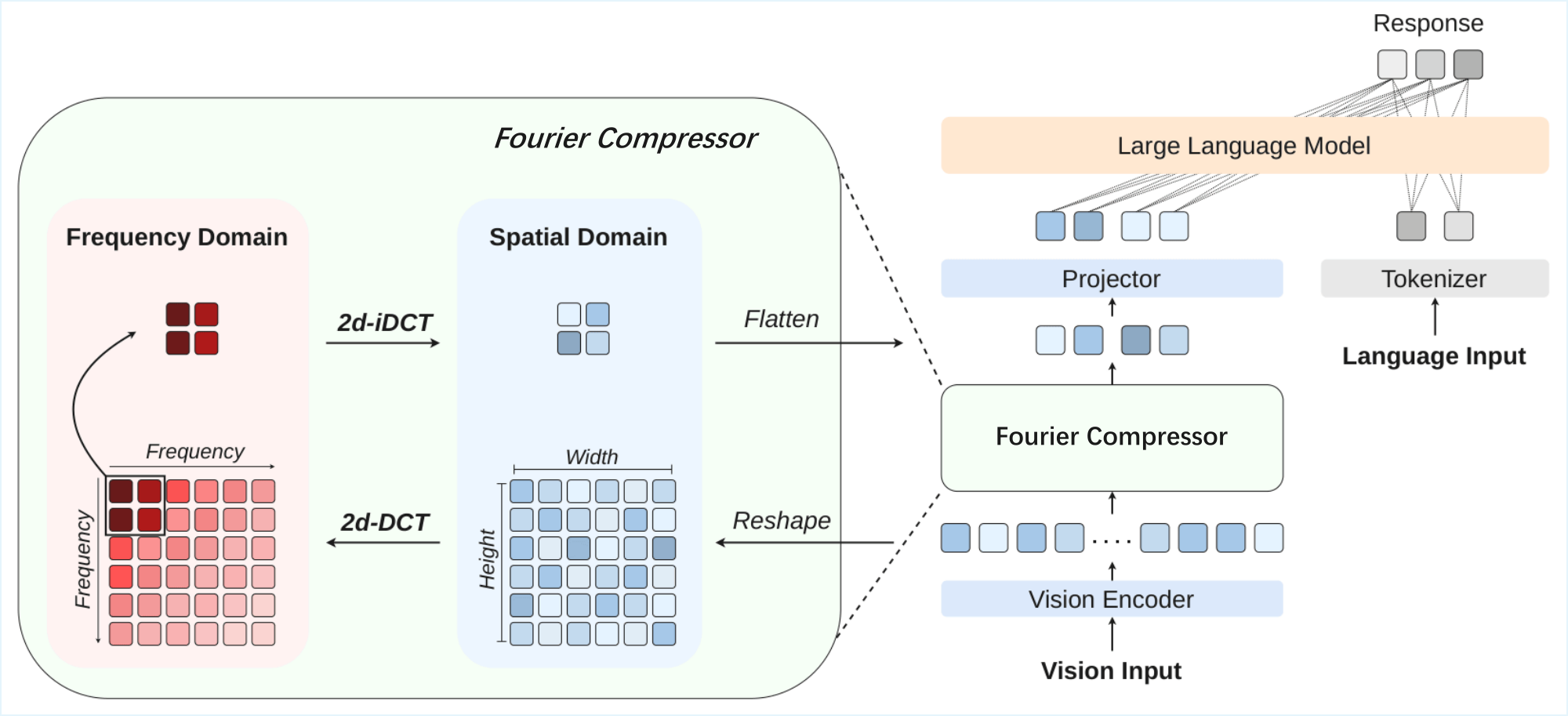}
    \caption{Illustration of Fourier Compressor. After passing through the vision encoder, visual features are reshaped into a grid and transformed into the frequency domain. Darker colors indicate larger frequency magnitudes, while lighter colors represent smaller magnitudes. Only the low-frequency components are retained and subsequently converted back to the spatial domain, serving as the compressed visual features.}
    \label{fig:framework}
\end{figure}

The overall architecture of our framework is illustrated in Figure~\ref{fig:framework}. The original vision input is first preprocessed (\eg, resizing or cropping for images, frame extraction for videos) before entering the vision encoder. The 3-channel resized image $\boldsymbol{X}^{v} \in \mathbb{R}^{r\times r\times 3}$, where $r$ represents the input resolution, is initially passed through a CNN to extract low-level features, resulting in grid image features $\hat{\boldsymbol{H}}^{v} \in \mathbb{R}^{N\times N\times h_{c}}$ where $N^2$ is the number of patches and $h_{c}$ is the output dimension of the CNN. These grid features are then flattened and passed through a pretrained ViT, which encodes the visual information into higher-level representations:
\begin{equation}
    \boldsymbol{H}^{v} = Vision\mbox{-}Encoder(\boldsymbol{X}^{v}),
\end{equation}
where $\boldsymbol{H}^{v} \in \mathbb{R}^{N^{2}\times h_{v}}$ and $h_{v}$ is the output dimension of the vision encoder.

Subsequently, the visual features are processed by our Fourier Compressor, which reduces the number of vision tokens from $N^2$ to $C^2$:
\begin{equation}
    \boldsymbol{H}^{v}_{c} = Fourier\mbox{-}Compressor(\boldsymbol{H}^{v}),
\end{equation}
where $\boldsymbol{H}^{v}_{c} \in \mathbb{R}^{C^{2}\times h_{v}}$.

Finally, the compressed visual features are passed through the projector to align with the text embedding space, and processed alongside the text instructions by the backbone LLM, following the typical VLM architecture.

\subsection{Fourier Compressor}
\label{subsec:module}

\begin{wrapfigure}[13]{r}{0.45\linewidth}
    \vspace{-2em}
    \begin{minipage}{\linewidth}
        \begin{algorithm}[H]
            \caption{Fourier Compressor}
            \label{alg:module}
            \footnotesize
            {\setlength{\baselineskip}{1.2\baselineskip}
            \begin{algorithmic}[1]
                \STATE {\bfseries Input:} Image feature $I$ of shape $(B, N^2, h_v)$
                \STATE {\bfseries Output:} Compressed $I^c$ of shape $(B, C^2, h_v)$
                \STATE $X \gets \text{Reshape}(\text{Transpose}(I, (1,2)), (B, h_v, N, N))$
                \STATE $F_1 \gets \text{DCT}(X)$
                \STATE $F_2 \gets \text{DCT}(\text{Transpose}(F_1, (2, 3)))$
                \STATE $F \gets \text{Transpose}(F_2, (2, 3))$
                \STATE $F^c \gets F[:, :, :C, :C]$
                \STATE $R_1 \gets \text{IDCT}(F^c)$
                \STATE $R_2 \gets \text{IDCT}(\text{Transpose}(R_1, (2, 3)))$
                \STATE $Y \gets \text{Transpose}(R_2, (2, 3))$
                \STATE $I^c \gets \text{Transpose}(\text{Reshape}(Y, (B, h_v, C^2)), (1,2))$
                \STATE {\bfseries return} $I^c$
            \end{algorithmic}
            }
        \end{algorithm}
    \end{minipage}
\end{wrapfigure}

Inside the Fourier Compressor, the original visual features are first reshaped back to the grid size:
\begin{equation}
    \boldsymbol{G}^{v} = Reshape(\boldsymbol{H}^{v}),
\end{equation}
where $\boldsymbol{G}^{v} \in \mathbb{R}^{N\times N\times h_{v}}$. Then, a two-dimensional Discrete Cosine Transform (2d-DCT) is applied along the two spatial dimensions of size $N$ to obtain the frequency representation $\hat{\boldsymbol{F}}^{v}$. Formally:
\begin{equation}
    \hat{\boldsymbol{F}}^{v}_{m, n, :} = \alpha_{m}\alpha_{n}\sum^{N-1}_{p=0} \sum^{N-1}_{q=0} \boldsymbol{G}^{v}_{p, q, :}\cdot \phi_{N}(m, p)\cdot \phi_{N}(n, q),
\end{equation}
where $\hat{\boldsymbol{F}}^{v} \in \mathbb{R}^{N\times N\times h_{v}}$ and $\phi_{\cdot}(\cdot, \cdot)$, $\alpha_{m}$ are specified in Equation~\ref{eq:DCT}. As analyzed in Section~\ref{subsec:analyses}, low-frequency components capture more global and semantically informative structures, which motivates the pruning of high-frequency components for compression:
\begin{equation}
    \boldsymbol{F}^{v} = \hat{\boldsymbol{F}}^{v}[0\colon \!\!C, 0\colon \!\!C, :],
\end{equation}
where $\boldsymbol{F}^{v} \in \mathbb{R}^{C\times C\times h_{v}}$ and $C^{2}$ is the number of preserved vision tokens.

Finally, we apply a two-dimensional inverse Discrete Cosine Transform (2d-iDCT) to reconstruct the spatial features from the frequency-domain tokens:
\begin{equation}
    \boldsymbol{T}^{v}_{i, j, :} = \sum^{C-1}_{p=0} \sum^{C-1}_{q=0} \alpha_{p}\alpha_{q}\cdot \boldsymbol{F}^{v}_{p, q, :}\cdot \phi_{C}(p, i)\cdot \phi_{C}(q, j),
\end{equation}
where $\boldsymbol{T}^{v} \in \mathbb{R}^{C\times C\times h_{v}}$, which is then flattened to obtain the compressed image features:
\begin{equation}
    \boldsymbol{H}^{v}_{c} = Flatten\left(\boldsymbol{T}^{v}\right),
\end{equation}
where $\boldsymbol{H}^{v}_{c} \in \mathbb{R}^{C^{2}\times h_{v}}$.

\subsection{Time Complexity}
\label{subsec:complexity}

A direct computation of the Discrete Cosine Transform (DCT) is inefficient, requiring $\mathcal{O}(N^2)$ operations for an N-point sequence. However, by leveraging the Fast Fourier Transform (FFT) operator, we can accelerate the DCT computation to $\mathcal{O}(N\log N)$, same for iDCT.

For a real-valued sequence of length N, denoted as $\langle x_{i}\rangle = \{x_{0}, x_{1}, \dots, x_{N-1}\}$, we first rearrange it by separating the even- and odd-indexed elements and reversing the order of the odd-indexed elements. If N is even, the reordered sequence $\langle y_{k} \rangle$ is defined as:
\begin{equation}
    y_{k} = 
    \begin{cases}
        x_{2k} & k=0,1,\dots,\frac{N-2}{2} \\
        x_{2N-1-2k} & k=\frac{N}{2},\dots,N-1
    \end{cases}
\end{equation}
That is, $\langle y_{k} \rangle = \{x_{0}, x_{2}, \dots, x_{N-2}, x_{N-1}, x_{N-3}, \dots, x_{1}\}$. A similar rearrangement applies when N is odd.

Next, we compute the FFT of $\langle y_{n} \rangle$, yielding the sequence $\langle z_{m} \rangle$:
\begin{equation}
        \langle z_{m}\rangle = FFT(\langle y_{n}\rangle),
\end{equation}
where $\langle z_{m}\rangle$ is a complex sequence of length N. The DCT coefficients can then be computed as:
\begin{equation}
        f_{k} = \Re(z_{k}) \cdot \cos\left(\frac{k\pi}{2N}\right) - \Im(z_{k}) \cdot \sin\left(\frac{k\pi}{2N}\right),
\end{equation}
where $\Re(\cdot)$ and $\Im(\cdot)$ denote the real and imaginary parts of a complex number, respectively.

For the two-dimensional DCT (2d-DCT) applied to a matrix $\boldsymbol{X} \in \mathbb{R}^{N\times N}$, the transformation is equivalent to performing a one-dimensional DCT along each row, followed by another one-dimensional DCT along each column. Consequently, applying 2d-DCT to an $N\times N$ matrix results in a total time complexity of $\mathcal{O}(N^2\log N)$.

Therefore, the Fourier Compressor, which first applies a 2d-DCT on the grid image features of size $N\times N\times h_{v}$ and then an 2d-iDCT on the compressed image features of size $C\times C\times h_{v}$, has an overall time complexity of:
\begin{equation}
    \mathbf{\mathcal{O}(N^2\log N + C^2\log C)}
\end{equation}

\begin{wraptable}[11]{r}{0.43\linewidth}
    \vspace{-1em}
    \caption{Time complexity of common modules processing inputs of shape $(B, N^2, h_v)$. $C^2$ denotes the number of learnable queries.}
    \label{tab:complexity}
    \centering
    \renewcommand{\arraystretch}{1.15}
    \footnotesize
    \begin{tabular}{ll}
        \toprule
        \textbf{Module} & \textbf{Time Complexity} \\
        \midrule
        MLP & $\mathcal{O}(B \cdot h_v^2 \cdot N^2)$ \\
        Self-Attention & $\mathcal{O}(B \cdot h_v \cdot N^4)$ \\
        Query Transformer & $\mathcal{O}(B \cdot h_v \cdot N^2 \cdot C^2)$ \\
        \bf Fourier Compressor & $\mathcal{O}(B \cdot h_v \cdot N^2 \log N)$ \\
        \bottomrule
    \end{tabular}
\end{wraptable}

Table~\ref{tab:complexity} summarizes the time complexity of representative modules operating on inputs of shape $(B, N^2, h_v)$. Under a typical regime where $h_v = \mathcal{O}(10^3)$, $N^2 = \mathcal{O}(10^2)$, and $C^2 = \mathcal{O}(10)$, our Fourier Compressor attains the lowest theoretical complexity, substantially outperforming attention-based (\eg, ATP-LLaVA~\citep{ATP-LLaVA}) and query-based (\eg, MQT-LLaVA~\citep{MQT-LLaVA}) compression methods in computational efficiency. Empirical evaluations further validate this efficiency advantage, as detailed in Section~\ref{subsec:efficiency}.

\section{Experiments}
\label{sec:experiments}

In this section, we first describe the training strategy and evaluation settings (Section~\ref{subsec:setup}). Then we present results on a wide range of image-based benchmarks (Section~\ref{subsec:image}) and demonstrate the generalizability of the Fourier Compressor across different VLM architectures (Section~\ref{subsec:generalization}), followed by empirical efficiency metrics (Section~\ref{subsec:efficiency}). Finally, we evaluate its applicability to video understanding tasks (Section~\ref{subsec:video}), highlighting its practical utility and broad applicability.

\subsection{Experimental Setup}
\label{subsec:setup}

Since frequency-domain truncation inevitably alters the feature distribution, additional training is required for the model to adapt to the compressed visual representations. Detailed training configurations are provided in Table~\ref{tab:training}.

\textbf{Fourier-LLaVA} incorporates the Fourier Compressor into the LLaVA-v1.5~\citep{LLaVA-v1.5} models. Training follows the original two-stage protocol of the base model: first aligning compressed visual features with the language embedding space, then optimizing for multimodal instruction following.

\begin{wraptable}[19]{r}{0.45\linewidth}
    \centering
    \setlength{\tabcolsep}{1.2mm}
    \renewcommand{\arraystretch}{1.15}
    \footnotesize
    \begin{tabular}{l|cc|c}
        \toprule
        \multirow{2}{*}{\bf Settings} & \multicolumn{2}{c|}{\bf Fourier-LLaVA} & \multirow{2}{*}{\bf Fourier-Qwen} \\
        & Stage 1 & Stage 2 & \\
        \midrule
        \multicolumn{4}{c}{\textit{\textcolor{gray}{Trainable modules}}} \\
        Vision Encoder &  &  & \checkmark \\
        Projector  & \checkmark & \checkmark & \checkmark \\
        LLM &  & \checkmark & \checkmark \\
        \midrule
        Dataset        & 558k   & 665k   & 600k \\
        Epochs         & 1      & 2      & 2 \\
        LoRA $r / \alpha$ & - & 128 / 256 & - \\
        Batch Size     & 256    & 256    & 128 \\
        Learning Rate  & 1e-3   & 2e-4   & 1e-6 \\
        MM LR          & -      & 2e-5   & 1e-5 \\
        Vision LR      & -      & -      & 1e-6 \\
        Optimizer      & \multicolumn{2}{c|}{AdamW} & AdamW \\
        Schedule       & \multicolumn{2}{c|}{Cosine} & Cosine \\
        Warmup Ratio   & \multicolumn{2}{c|}{0.03} & 0.03 \\
        \bottomrule
    \end{tabular}
    \caption{Training configurations for Fourier-LLaVA and Fourier-Qwen.}
    \label{tab:training}
\end{wraptable}

\textbf{Fourier-Qwen} incorporates the Fourier Compressor into the Qwen-VL series~\citep{qwen2vl,qwen25vl}. We perform continued finetuning on 600k single-image conversation samples from LLaVA-NeXT~\citep{LLaVA-v1.6}, inserting the compressor after the original MLP-based merger.

We evaluate on eight image-based benchmarks covering VQA, domain knowledge, and text recognition~\citep{vqav2,GQA,SciQA,textvqa,pope,mmbench,LLaVA-v1,mmmu}. For the Fourier-Qwen series, three additional benchmarks are included~\citep{mme,realworldqa,mmstar}. We primarily use \textit{lmms-eval}~\citep{lmms-eval} for evaluation. For benchmarks not supported by lmms-eval ($\text{VQA}^{\text{T}}$, MMB, and $\text{LLaVA}^{\text{W}}$), we follow the official scripts from LLaVA-v1.5. For $\text{LLaVA}^{\text{W}}$, performance is measured with \textit{gpt-4-0613} at temperature 0.2.

\subsection{Image Benchmark Performance}
\label{subsec:image}

\begin{table}[t]
    \caption{Performance of Fourier-LLaVA on 8 image-based benchmarks. ``\#I" denotes the number of vision tokens per image. All baseline results are reported from original papers, except MQT-LLaVA on $\text{VQA}^{\text{T}}$, which was not originally reported and is newly evaluated by us. Note that PruMerge is the only baseline available in the 13B model setting.}
    \label{tab:img}
    \centering
    \setlength{\tabcolsep}{2mm}
    \renewcommand{\arraystretch}{1.15}
    \footnotesize
    \begin{tabular}{l|c|cccccccc|c}
        \toprule
        \bf Model & \bf \#I & $\textbf{VQA}^{\text{v2}}$ & \bf GQA & \bf SQA & $\textbf{VQA}^{\text{T}}$ & \bf POPE & \bf MMB & $\textbf{LLaVA}^{\text{W}}$ & \bf MMMU & \bf Avg. \\
        \midrule
        \midrule
        LLaVA-v1.5-7B & 576 & 78.5 & 62.0 & 66.8 & 58.2 & 85.9 & 64.3 & 65.4 & 35.3 & 64.6 \\
        \midrule
        \multicolumn{11}{c}{\textit{\textcolor{gray}{Compression ratio: 55.6\%}}} \\
        MQT-LLaVA~\citep{MQT-LLaVA} & 256 & 76.8 & 61.6 & 67.6 & 53.2 & 84.4 & 64.3 & \bf 64.6 & \bf 34.8 & 63.4 \\
        \textbf{Fourier-LLaVA} & 256 & \bf 78.6 & \bf 62.7 & \bf 69.9 & \bf 56.0 & \bf 85.3 & \bf 66.4 & 64.4 & 33.1 & \bf 64.6 \\
        \midrule
        \multicolumn{11}{c}{\textit{\textcolor{gray}{Compression ratio: 75.0\%}}} \\
        ATP-LLaVA~\citep{ATP-LLaVA} & 144 & 76.4 & 59.5 & \bf 69.1 & - & 84.2 & \bf 66.0 & - & - & - \\
        MQT-LLaVA~\citep{MQT-LLaVA} & 144 & 76.4 & 61.4 & 67.5 & 52.6 & 83.9 & 64.4 & 61.4 & 34.4 & 62.8 \\
        Prumerge~\citep{LLaVA-PruMerge} & 144 & 76.8 & - & 68.3 & \bf 57.1 & 84.0 & 64.9 & - & - & -  \\
        \textbf{Fourier-LLaVA} & 144 & \bf 77.7 & \bf 61.7 & 69.0 & 54.7 & \bf 85.0 & 65.6 & \bf 67.8 & \bf 35.3 & \bf 64.6  \\
        \midrule
        \multicolumn{11}{c}{\textit{\textcolor{gray}{Compression ratio: 88.9\%}}} \\
        ATP-LLaVA~\citep{ATP-LLaVA} & 88 & 73.3 & 56.8 & 67.2 & - & 82.6 & \bf 64.7 & - & - & -  \\
        MQT-LLaVA~\citep{MQT-LLaVA} & 64 & 75.3 & 60.0 & 67.0 & 51.7 & 83.6 & 63.5 & 59.4 & \bf 34.4 & 61.9  \\
        \textbf{Fourier-LLaVA} & 64 & \bf 76.3 & \bf 60.4 & \bf 69.3 & \bf 52.6 & \bf 85.3 & \bf 64.7 & \bf 63.1 & \bf 34.4 & \bf 63.3 \\
        \midrule
        \multicolumn{11}{c}{\textit{\textcolor{gray}{Compression ratio: 93.75\%}}} \\
        MQT-LLaVA~\citep{MQT-LLaVA} & 36 & 73.7 & 58.8 & 66.8 & 50.4 & 81.9 & 63.4 & 59.6 & \bf 34.4 & 61.1  \\
        PruMerge~\citep{LLaVA-PruMerge} & 32 & 72.0 & - & 68.5 & \bf 56.0 & 76.3 & 60.9 & - & - & -  \\
        \textbf{Fourier-LLaVA} & 36 & \bf 74.9 & \bf 59.5 & \bf 69.0 & 51.0 & \bf 84.0 & \bf 64.3 & \bf 61.1 & 32.8 & 
        \bf 62.1  \\
        \midrule
        \midrule
        LLaVA-v1.5-13B & 576 & 80.0 & 63.3 & 71.6 & 61.3 & 85.9 & 67.7 & 72.5 & 36.4 & 67.3 \\
        \midrule
        \multicolumn{11}{c}{\textit{\textcolor{gray}{Compression ratio: 75.0\%}}} \\
        PruMerge~\citep{LLaVA-PruMerge} & 144 & 77.8 & - & 71.0 & \bf 58.6 & 84.4 & 65.7 & - & - & - \\
        \textbf{Fourier-LLaVA} & 144 & \bf 78.7 & 62.7 & \bf 71.1 & 57.4 & \bf 85.4 & \bf 66.2 & 69.5 & 35.8 & 65.9 \\
        \bottomrule
    \end{tabular}
\end{table}

We begin by evaluating the effectiveness of the proposed Fourier Compressor on image-based multimodal benchmarks. Specifically, we integrate the compressor into the LLaVA-v1.5 backbone and systematically vary the number of preserved vision tokens from 36 to 256.

Table~\ref{tab:img} presents the results on eight image benchmarks. Notably, even with a reduced number of vision tokens, Fourier-LLaVA achieves competitive results. With 256 vision tokens (44\% of the original tokens), Fourier-LLaVA achieves superior performance to the vanilla LLaVA-v1.5-7B on four benchmarks. Moreover, with just 144 vision tokens ($25\%\times 576$), it still surpasses the base model in terms of average score. Even at the lowest setting of 36 vision tokens (6.25\% of the original tokens), Fourier-LLaVA only exhibits a 3.87\% drop in average, while continuing to outperform the base model on SciQA and MMB.

For other token compression approaches, we compare the performance under the same number of vision tokens. Note that all the methods mentioned are implemented based on LLaVA-v1.5 series and require a two-stage training process. Generally, Fourier-LLaVA outperforms prior parameter-free methods such as ATP-LLaVA and LLaVA-PruMerge across nearly all benchmarks at the same vision token count, substantially improving generation performance under high compression ratios. Furthermore, it surpasses MQT-LLaVA, which relies on additional learnable parameters, achieving a 2.2\% increase in average scores, further demonstrating the effectiveness of our approach.

Moreover, when scaled to a 13B backbone, Fourier-LLaVA continues to outperform LLaVA-PruMerge under the same vision token budget. Besides, with only 25\% of the original vision tokens, our approach incurs only a 2.1\% drop in the average score compared to the base model.

Overall, these results demonstrate that our Fourier Compressor achieves superior performance when compressing visual tokens for image inputs, maintaining strong robustness under aggressive token compression. Besides, it also exhibits strong compression efficiency, as detailed in Subsection~\ref{subsec:efficiency}.

\subsection{Generalization Across Architectures}
\label{subsec:generalization}

\begin{table}[t]
    \caption{Performance of Fourier-Qwen series. ``\#I" denotes the average number of vision tokens per image. We report the performance of all models using \textit{lmms-eval}, with the number of visual tokens standardized to 256 - 2304 for fair comparison.}
    \label{tab:qwen}
    \centering
    \renewcommand{\arraystretch}{1.15}
    \footnotesize
    \begin{tabular}{cl|cc|cccccc}
        \toprule
        \multicolumn{2}{c|}{\bf Model Series} & \bf \#I & \bf Avg. & \bf MME & $\textbf{VQA}^{\text{T}}$ & \bf POPE & \bf RWQA & \bf MMB & \bf MMS\\
        \midrule
        \midrule
        \multirow{3}{*}{\bf Qwen2-VL-2B} & Vanilla & 553 & 66.8 & 1894.4 & 78.2 & 86.1 & 61.4 & 72.3 & 44.2 \\
        & \bf Fourier- & 236 & 65.7 & 1917.1 & 75.4 & 86.5 & 61.6 & 72.6 & 43.3 \\
        &  & \it -57.3\% & \it -1.6\% & \it +1.2\% & \it -3.6\% & \it +0.5\% & \it +0.3\% & \it +0.4\% & \it -2.0\% \\
        \midrule
        \multirow{3}{*}{\bf Qwen2.5-VL-3B} & Vanilla & 553 & 71.1 & 2138.3 & 77.6 & 87.2 & 59.6 & 77.8 & 56.2 \\
        & \bf Fourier- & 236 & 72.6 & 2110.8 & 76.1 & 87.8 & 64.3 & 76.8 & 55.3 \\
        &  & \it -57.3\% & \it +2.1\% & \it -1.3\% & \it -1.9\% & \it +0.7\% & \it +7.9\% & \it -1.3\% & \it -1.6\% \\
        \bottomrule
    \end{tabular}
\end{table}

To assess the generalizability of the Fourier Compressor beyond the LLaVA series, we further apply it to Qwen2-VL-2B and Qwen2.5-VL-3B~\citep{qwen2vl, qwen25vl}, and evaluate on six image-based benchmarks. This evaluation examines whether frequency-domain compression remains effective across different vision encoders, backbone LLMs, and input resolutions.

As shown in Table~\ref{tab:qwen}, our method consistently preserves strong performance under substantial token reduction. With a 57.3\% decrease in vision tokens, Fourier-Qwen2-VL-2B incurs only a marginal 1.6\% drop in average performance, while even achieving improvements on MME, POPE, RWQA, and MMB. Notably, on Qwen2.5-VL-3B, the compressed model surpasses the vanilla baseline by 2.1\% on average, demonstrating that token reduction does not necessarily compromise reasoning capability. In particular, the 7.9\% gain on RealWorldQA suggests that frequency-domain compression may help suppress redundant or noisy visual components, benefiting real-world question answering tasks.

Overall, these results indicate that the effectiveness of the Fourier Compressor is not architecture-specific. It generalizes robustly across model scales and input resolutions, maintaining or even improving performance despite aggressive visual token compression.

\subsection{Floating Point Operations and Latency}
\label{subsec:efficiency}

\begin{figure}[!t]
    \centering
    \includegraphics[width=0.7\linewidth]{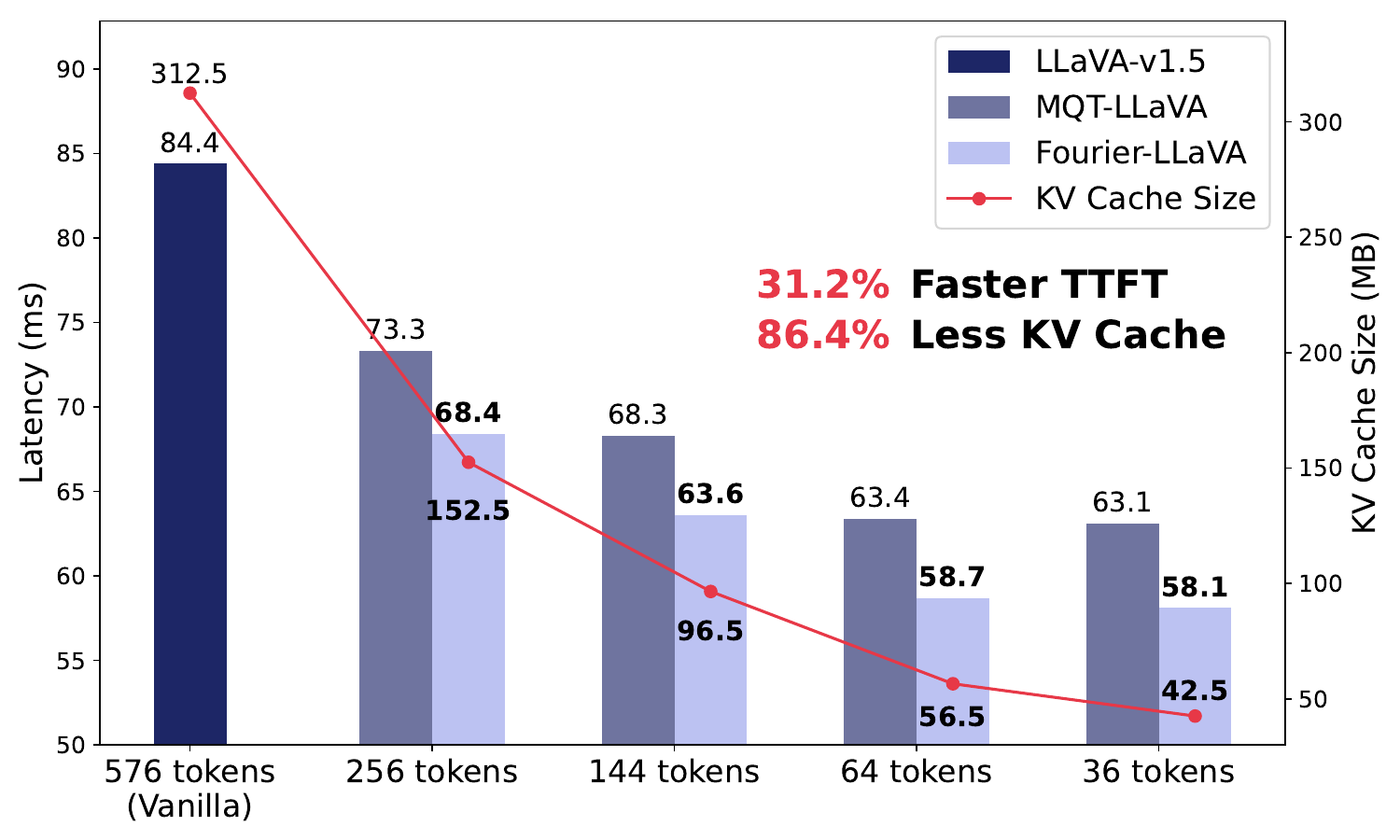}
    \caption{Latency and KV cache usage of Fourier-LLaVA under varying numbers of vision tokens.}
    \label{fig:latency}
\end{figure}

\begin{wraptable}[11]{r}{0.4\linewidth}
    \vspace{-1em}
    \caption{Floating Point Operations of Fourier-LLaVA under varying numbers of preserved vision tokens (\#I).}
    \label{tab:flops}
    \centering
    \renewcommand{\arraystretch}{1.15}
    \footnotesize
    \begin{tabular}{lc|cc}
        \toprule
        \bf Model & \bf \#I & \bf FLOPs (T) & \bf ↓ (\%) \\
        \midrule
        LLaVA-v1.5-7B & 576 & 8.54 & - \\
        \midrule
        \multirow{4}{*}{\bf Fourier-LLaVA} & 256 & 4.30 & 49.6 \\
        & 144 & 2.81 & 67.1 \\
        & 64 & 1.75 & 79.5 \\
        & 36 & 1.38 & 83.8 \\
        \bottomrule
    \end{tabular}
\end{wraptable}

Beyond effectiveness, a key advantage of the proposed Fourier Compressor lies in its parameter-free design and inherent computational efficiency due to its FFT-based implementation, as discussed in Section~\ref{subsec:complexity}. To empirically validate that these theoretical advantages translate into practical gains, we conduct a systematic evaluation of the efficiency of Fourier Compressor.

We evaluate the FLOPs (Floating Point Operations) of Fourier-LLaVA under varying numbers of vision tokens using \textit{calflops} on an RTX 4090 (24G) GPU. Additionally, we measure the TTFT (Time to First Token) and KV cache usage on an A100 (40G) GPU.

As illustrated in Table~\ref{tab:flops} and Figure~\ref{fig:latency}, Fourier-LLaVA achieves a substantial reduction in computational cost compared to LLaVA-v1.5-7B, reducing FLOPs by 83.8\%. Furthermore, our method lowers KV cache usage by 86.4\% and improves inference speed by 31.2\% in TTFT, outperforming other token compression baselines such as MQT-LLaVA under equivalent visual token counts.

These improvements highlight the efficiency of our parameter-free approach, distinguishing Fourier Compressor from prior works that rely on attention-based reductions or learnable parameters, which is crucial for enabling real-time deployment of VLMs on resource-constrained devices.

\subsection{Applicability to Video Tasks }
\label{subsec:video}

\begin{table}[t]
    \caption{Performance of Fourier-LLaVA and Fourier-Qwen on MVBench. ``\#V" denotes the average number of visual tokens per video. We report the performance of baselines and our models using lmms-eval. For fair comparison, when evaluating Fourier-Qwen series, the number of visual tokens per frame is set to 256 - 384, and the maximum number of frames is 16.}
    \label{tab:mvbench}
    \centering
    \renewcommand{\arraystretch}{1.15}
    \footnotesize
    \begin{tabular}{lc|c>{\centering\arraybackslash}p{0.75cm}|ccccccccc}
        \toprule
        \bf Model & \bf Size & \bf \#V & \bf Avg. & \bf Ac & \bf Ob & \bf Posi & \bf Sc & \bf Cou & \bf At & \bf Pose & \bf Ch & \bf Cog \\
        \midrule
        \midrule
        VideoChatGPT & 7B & 356 & 32.7 & 32.1 & 40.7 & 21.5 & 31.0 & 28.0 & 44.0 & 29.0 & 33.0 &  30.3 \\
        Video-LLaMA & 7B & 40 & 34.1 & 34.4 & 42.2 & 22.5 & 43.0 & 28.3 & 39.0 & 32.5 & 40.0 & 29.3 \\
        Video-LLaVA & 7B & 4096 & 43.1 & 48.0 & 46.5 & 27.8 & 84.5 & 35.5 & 45.8 & 34.0 & 42.5 & 34.2 \\
        VideoChat & 7B & 4096 & 35.5 & 38.0 & 41.2 & 26.3 & 48.5 & 27.8 & 44.3 & 26.5 & 41.0 & 27.7 \\
        VideoChat2 & 7B & 4096 & 51.1 & 61.3 & 57.3 & 23.0 & 88.5 & 40.5 & 51.3 & 49.0 & 36.5 & 47.0 \\
        \midrule
        LLaVA-v1.5 & 7B & 2304 & 45.6 & 52.8 & 43.7 & 31.5 & 83.5 & 37.0 & 45.3 & 44.5 & 50.0 & 37.3 \\
        \bf Fourier-LLaVA & 7B & 288 & 44.0 & 48.1 & 46.3 & 31.0 & 81.0 & 40.5 & 41.3 & 36.0 & 49.0 & 36.3 \\
        \quad \textit{$\Delta$ vs. LLaVA-v1.5} &  & \it -87.5\% & \it -1.6 & \it -4.7 & \it +2.6 & \it -0.5 & \it -2.5 & \it +3.5 & \it -4.0 & \it -8.5 & \it -1.0 & \it -1.0 \\
        \midrule
        Qwen-2-VL & 2B & 3193 & 61.4 & 68.5 & 65.7 & 44.8 & 90.5 & 60.8 & 65.3 & 53.5 & 59.0 & 47.8 \\
        \bf Fourier-Qwen-2 & 2B & 1328 & 59.8 & 67.7 & 63.3 & 40.0 & 88.5 & 55.5 & 64.8 & 51.0 & 57.0 & 50.2 \\
        \quad \textit{$\Delta$ vs. Qwen-2-VL} &  & \it -58.4\% & \it -1.6 & \it -0.8 & \it -2.4 & \it -4.8 & \it -2.0 & \it -5.3 & \it -0.5 & \it -2.5 & \it -2.0 & \it +2.4 \\
        \midrule
        Qwen-2.5-VL & 3B & 3193 & 65.2 & 68.3 & 67.7 & 48.8 & 91.0 & 63.0 & 75.0 & 51.5 & 75.5 & 56.0 \\
        \bf Fourier-Qwen-2.5 & 3B & 1328 & 62.9 & 67.2 & 65.7 & 48.5 & 90.5 & 54.8 & 70.5 & 47.5 & 66.0 & 57.5 \\
        \quad \textit{$\Delta$ vs. Qwen-2.5-VL} &  & \it -58.4\% & \it -2.3 & \it -1.1 & \it -2.0 & \it -0.3 & \it -0.5 & \it -8.2 & \it -4.5 & \it -4.0 & \it -9.5 & \it +1.5 \\
        \bottomrule
    \end{tabular}
\end{table}

Although our method is primarily designed for visual token compression in image inputs, the proposed Fourier Compressor is inherently model-agnostic and parameter-free, enabling zero-shot application to video tasks. To demonstrate this extension capability, we evaluate both Fourier-LLaVA and Fourier-Qwen models on MVBench~\citep{MVBench}, a comprehensive multi-modal video understanding benchmark comprising 20 challenging tasks.

As shown in Table~\ref{tab:mvbench}, our method leverages significantly fewer vision tokens while outperforming various open-source VLMs. Specifically, Fourier-Qwen series achieves a 58.4\% reduction in visual tokens with only a 3.1\% drop in average performance, while Fourier-LLaVA retains 96.5\% of the base model's performance utilizing merely 12.5\% of the video tokens, and still surpasses the base model on the Object and Count tasks. This zero-shot capability in video understanding highlights the efficiency and robustness of our method, and suggests promising directions for future work on scaling frequency-aware compression to visual representations of video inputs.

\section{Conclusion}
\label{sec:conclusion}

In this paper, we propose \textbf{Fourier Compressor}, an efficient, effective, and highly generalizable framework for vision token compression in Vision-Language Models. Motivated by the non-uniform frequency distribution of semantic information, our method significantly reduces frequency-domain redundancy from visual representations. Empirically, it achieves a favorable trade-off between performance and efficiency: Without introducing any additional parameters, it preserves over 96\% average accuracy across eight image-based benchmarks while retaining only 6.25\% of the original visual tokens, leading to a substantial reduction in FLOPs to 16.16\% and an inference speedup of 31.2\%. Seamlessly integrated into both the LLaVA-v1.5 and Qwen-VL series, Fourier Compressor exhibits strong generalization across model architectures, and further demonstrates promising performance on video understanding tasks. These results collectively highlight a favorable trade-off between computational cost and model performance, facilitating more efficient and scalable deployment of VLMs in real-world scenarios.

\bibliography{main}

\end{document}